\documentclass{article}

\usepackage[final]{colm2026_conference}

\usepackage{microtype}

\usepackage{hyperref}
\usepackage{url}
\usepackage{lineno}

\definecolor{darkblue}{rgb}{0, 0, 0.5}
\hypersetup{colorlinks=true, citecolor=darkblue, linkcolor=darkblue, urlcolor=darkblue}

\usepackage{graphicx}
\usepackage{subcaption}
\usepackage{booktabs} 

\usepackage{amsmath}
\usepackage{amssymb}
\usepackage{mathtools}
\usepackage{amsthm}
\usepackage{multirow}
\usepackage{enumitem}
\usepackage{float}
\usepackage{wrapfig}

\usepackage[capitalize,noabbrev]{cleveref}

\usepackage{xcolor, adjustbox}

\definecolor{posgreen}{HTML}{2E7D32}
\definecolor{negred}{HTML}{C62828}
\definecolor{outblue}{HTML}{1565C0}

\usepackage{tcolorbox}
\tcbset{
    left*=10pt, right*=10pt,
    top=4pt, bottom=4pt,
    colback=white,
    colframe=black!75,
    fonttitle=\bfseries,
}

\theoremstyle{plain}

\theoremstyle{definition}

\theoremstyle{remark}

\usepackage{algorithm}
\usepackage{algorithmic}
\renewcommand{\algorithmiccomment}[1]{%
  \hfill\textcolor{blue!40}{\(\triangleright\) #1}%
}

\newcommand{\methodname}{\texttt{ROTATE}}

\newcommand{\ww}{$\mathbf{w}$}

\definecolor{mygreen}{HTML}{7ecc7b}
\definecolor{mygray}{HTML}{8c8c8c}
\definecolor{mywingreen}{HTML}{4daf50}
\definecolor{mywingray}{HTML}{bdbdbd}
\definecolor{mywinred}{HTML}{e93636}

\title{Disentangling MLP Neuron Weights in Vocabulary Space}

\author{Asaf Avrahamy \quad
  Yoav Gur-Arieh \quad
Mor Geva \vspace{5pt}
\\
Blavatnik School of Computer Science and AI, Tel Aviv University \vspace{3pt} \\ \texttt{\{asafavrahamy@mail, yoavgurarieh@mail, morgeva@tauex\}.tau.ac.il}
}

\begin{document}
\ifcolmsubmission
\linenumbers
\fi
\maketitle

\begin{abstract}
Interpreting the information encoded in language model weights remains a fundamental challenge in mechanistic interpretability.
In this work, we introduce \methodname{} (Rotation-Optimized Token Alignment in weighT spacE), a data-free method requiring no forward passes that disentangles MLP neurons directly in weight space. Our approach relies on a key statistical observation: neurons that encode coherent, monosemantic concepts exhibit high kurtosis when projected onto the model's vocabulary. By optimizing rotations of neuron weights to maximize their vocabulary-space kurtosis, our method recovers sparse, interpretable directions which we name \textit{vocabulary channels}.
Experiments on Llama-3.1-8B-Instruct and Gemma-2-2B-it demonstrate that \methodname{} consistently recovers vocabulary channels that are faithful to the neuron's behavior; ablating individual channels selectively disables corresponding input activations or the promotion of specific concepts.
Moreover, aggregating channel-level descriptions yields comprehensive neuron descriptions that outperform optimized activation-based baselines by 2–3× in head-to-head comparisons. By providing a data-free decomposition of neuron weights, \methodname{} offers a scalable, fine-grained building block for interpreting language models.
\end{abstract}

\section{Introduction}

One of the underexplored goals of mechanistic interpretability is inspecting the information encoded in language model (LM) weights.
Targeting weights is particularly appealing as it allows examining the model independently of specific inputs or data distributions, which can introduce biases \citep{interp-illus, gao2025scaling} or incur high computational costs.
A key challenge in interpreting LM weights is finding the ``right unit of analysis'' \citep{mueller2025quest, sharkey2025open, geiger2025causal}. While prior work has made progress in identifying neurons that capture individual, coherent concepts \citep{kv-memory, ff-pred, dai-know} and attention heads that implement specific functions \citep{attention-survey, maps}, in most cases these components are polysemantic and encode multiple entangled concepts \citep{interp-illus, haystack}.

\begin{figure}[t]
\setlength{\belowcaptionskip}{-10pt}
    \centering
    \includegraphics[width=1.0\linewidth]{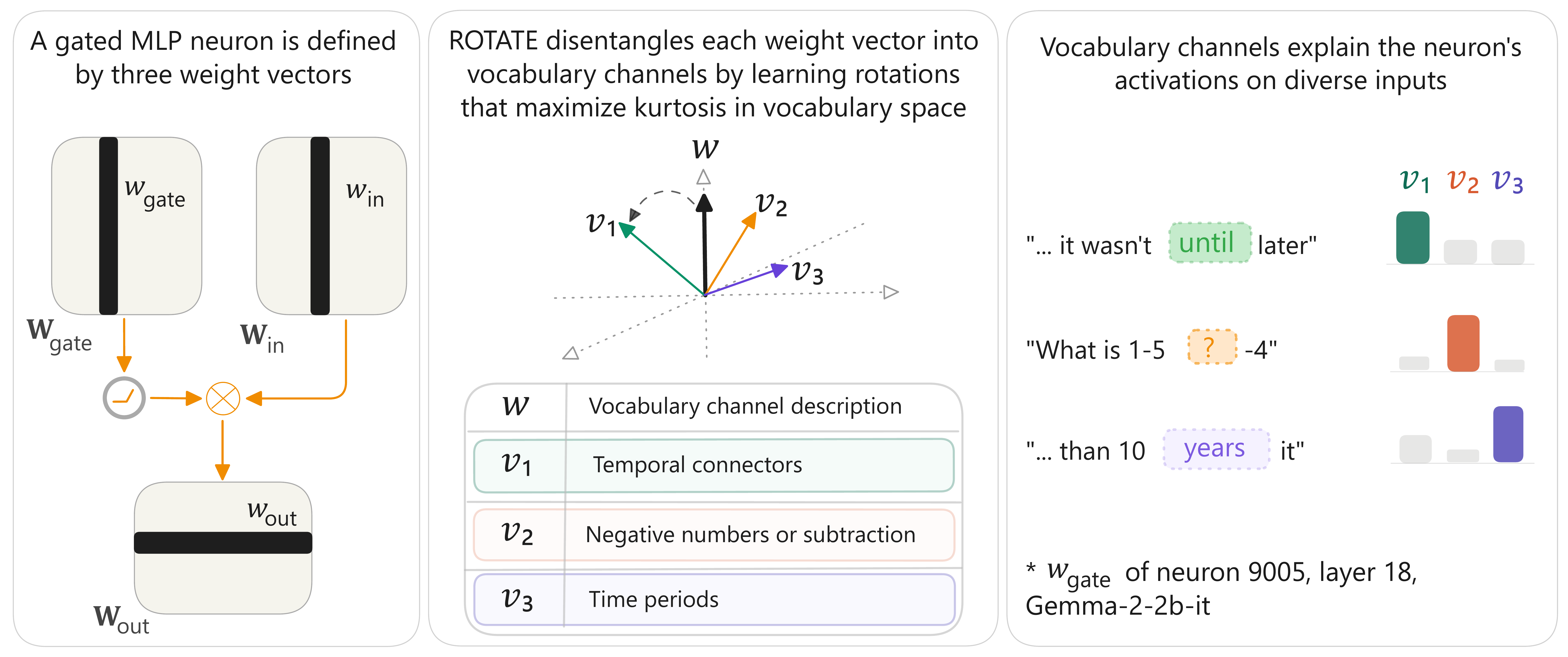}
    \caption{We propose to disentangle MLP neuron weights (\textbf{Left}) 
    using \methodname{}, a data-free method that learns rotations of a neuron's weight vector $\mathbf{w}$ to maximize kurtosis in the model's vocabulary space, recovering sparse, interpretable directions we call \emph{vocabulary channels} (\textbf{Middle}). 
    Each channel isolates a distinct concept encoded in $\mathbf{w}$, allowing a fine-grained understanding of the neuron's mechanism across diverse inputs (\textbf{Right}).}
    \label{fig:intro}
\end{figure}

In this work, we tackle the challenge of polysemanticity by disentangling model weights, focusing on MLP neurons in LMs. First, we make a key observation: MLP neurons that strongly promote single, coherent concepts exhibit high kurtosis when their weights are projected into the model's vocabulary space. This suggests that kurtosis in vocabulary space—a measure of how heavy-tailed the distribution over vocabulary tokens is—can serve as a proxy for directions with monosemantic attributes.
Based on this observation, we introduce \methodname{} (Rotation-Optimized Token Alignment in weighT spacE), a data-free method requiring no forward passes through the model that disentangles MLP neuron weights into their constituent, human-interpretable components.
Given a neuron weight vector $ \mathbf{w} \in \mathbb{R}^{d}$, \methodname{} learns rotation matrices \{$\mathbf{R_i}$\}, each rotating \ww{} to reveal a semantically privileged basis in weight space $\mathbf{v}_i := \mathbf{R}_i \mathbf{w}$ (see Figure~\ref{fig:intro}). Rotations are learned by optimizing towards increased vocabulary space kurtosis, while penalizing deviations from \ww{}. We call these discovered vectors $\{\mathbf{v}_i\}$ \textit{vocabulary channels}, as they are projections of the original neuron that are aligned with the vocabulary basis of the model.

Through a series of experiments on Gemma-2-2B-it \citep{gemma2} and Llama-3.1-8B-Instruct \citep{llama3}, we show that vocabulary channels capture fine-grained functions that are faithful to the neuron's behaviors. Ablating individual channels selectively suppresses specific neuron functionalities without affecting others. Moreover, vocabulary channels provide more complete neuron explanations, covering a wider range of the neuron's activation space.
Across both these evaluations, \methodname{} outperforms decompositions by state-of-the-art sparse autoencoders (SAEs), Gemma Scope \citep{gemmascope} and Llama Scope \citep{llamascope}, applied to neuron weights.
Next, we demonstrate the utility of \methodname{} in generating natural-language neuron descriptions. By aggregating the descriptions of a neuron's channels, we produce descriptions that consistently outperform optimized descriptions over top-activating inputs \citep{transluce} and a strong baseline that combines activating inputs with vocabulary projection \citep{output-enhancing}, 
achieving 2–3× higher win rates in head-to-head comparisons across layers and evaluation sets.

In summary, our work makes the following contributions: 
(a) we observe that high-kurtosis vocabulary distributions correlate with monosemantic directions in LM weight space,
(b) we introduce \methodname{}, a data-free method that uses this signal for disentangling MLP weights into interpretable directions, 
(c) experiments on widely-used LMs show that \methodname{} recovers faithful vocabulary channels that outperform SAE-based baselines on both faithfulness to neuron behavior and coverage of its activation spectrum, and (d) we show that aggregating vocabulary channels can produce better neuron descriptions than common automated interpretability approaches.
We release our code at \url{https://github.com/AsafAvr/rotating-neurons}.

\section{Preliminaries and notation}
\label{sec:preliminaries}

\paragraph{Neurons in LMs with gated MLP layers} 
We focus on autoregressive transformer-based \citep{attention} LMs with a hidden dimension $d$ and an inner MLP dimension $d_a$. Let $\mathbf{E} \in \mathbb{R}^{V \times d}$ and $\mathbf{U} \in \mathbb{R}^{d \times V}$ denote the embedding and unembedding matrices, where $V$ is the vocabulary size.
A gated MLP layer \citep{gatedmlps} is defined by three parameter matrices $\mathbf{W}_{\text{gate}}, \mathbf{W}_{\text{in}}, \mathbf{W}_{\text{out}}^T \in \mathbb{R}^{d_{a} \times d}$ and a nonlinear activation function $\sigma$:\footnote{Our approach also can be applied to vanilla MLPs with only $\mathbf{W}_{\text{in}}$ and $\mathbf{W}_{\text{out}}$.}
\begin{equation}
    \text{MLP}(\mathbf{x}) = \mathbf{W}_{\text{out}} \left( \sigma(\mathbf{W}_{\text{gate}} \mathbf{x}) \odot (\mathbf{W}_{\text{in}} \mathbf{x}) \right)
\label{eq:mlp_block}
\end{equation}
where $\mathbf{x} \in \mathbb{R}^{d}$ is an input hidden state and $\odot$ denotes element-wise multiplication. 
A neuron is defined by an index $i \in [d_{a}]$ and acts as a computational unit with three weight vectors:
\textbf{Input vectors} $\mathbf{w}_{\text{gate}}^{(i)}, \mathbf{w}_{\text{in}}^{(i)} \in \mathbb{R}^d$, which correspond to the $i$-th rows of $\mathbf{W}_{\text{gate}}$ and $\mathbf{W}_{\text{in}}$, respectively, and an \textbf{output vector} $\mathbf{w}_{\text{out}}^{(i)} \in \mathbb{R}^d$, corresponding to the $i$-th column of $\mathbf{W}_{\text{out}}$. The input vectors determine the neuron's activation pattern for a given input $\mathbf{x}$, while the output vector is written to the residual stream, weighted by the input's activation strength.

\paragraph{Vocabulary projection}
Projection to vocabulary space has been a common approach for analyzing model representations and weights \citep{logitlens,ff-pred,vocab-proj}. The projection $\mathbf{z} = \mathbf{w} \mathbf{U}$ of a neuron's weight vector $\mathbf{w}$ yields a vector of logits $\mathbf{z} \in \mathbb{R}^V$, where the indices of the highest and lowest values in $\mathbf{z}$ correspond to the tokens that the neuron most strongly promotes or suppresses, respectively.

\paragraph{Kurtosis}
Kurtosis is the fourth standardized moment, which provides a statistical measure of the ``tailedness'' of a probability distribution. 
Here, we treat the logits $\mathbf{z} \in \mathbb{R}^V$ as a distribution over the vocabulary. A high kurtosis value indicates that the distribution is sharply peaked with heavy tails, meaning the neuron acts strongly on a sparse set of tokens while having little effect on most others. Thus, Gaussianity represents the ``least interesting'' distribution, and we maximize kurtosis to identify directions that are non-Gaussian, separating mixed signals into independent, sparse components. 
For the definition of kurtosis and an illustration, see \S\ref{appx:add-pre}.

\section{High vocabulary kurtosis as a signal of monosemantic directions}
\label{sec:kurtosis}

To disentangle polysemantic neurons in weight space without ground-truth labels, we require an unsupervised measure that distinguishes interpretable, concept-centric directions from entangled or random ones. 
In this section, we identify vocabulary-projection kurtosis (vocabulary kurtosis in short), as such a signal. We ground this hypothesis with observations from prior work and validate it through empirical analysis.

\paragraph{Monosemantic neurons in LMs}
Prior work has identified neurons in LMs that strongly encode single, coherent concepts. \citet{ff-pred} showed that 
neuron weight vectors in $\mathbf{W}_{\text{out}}$ can be viewed as additive updates that promote the probability of a sparse set of semantically related tokens.
More recently, \citet{universal, remarkable} identified a small set of ``universal'' neurons, characterized by high kurtosis in the vocabulary basis, that cluster densely in the middle-to-late layers during the ``prediction ensembling'' stage, suggesting that sparse, heavy-tailed distributions are a signature of output-facing computations.
Last, \citet{conceptvec} found a set of MLP neurons called \textit{concept vectors} in Llama-2-7B \citep{llama2} and OLMo-7B \citep{olmo}, that exhibit monosemantic patterns in their vocabulary projections. These neurons strongly promote specific concepts, and ablating them degrades the model's ability to generate knowledge about the concepts they encode. 

\begin{wrapfigure}{r}{0.45\columnwidth}
    \vspace{-12pt}
    \centering
    \includegraphics[width=0.43\columnwidth]{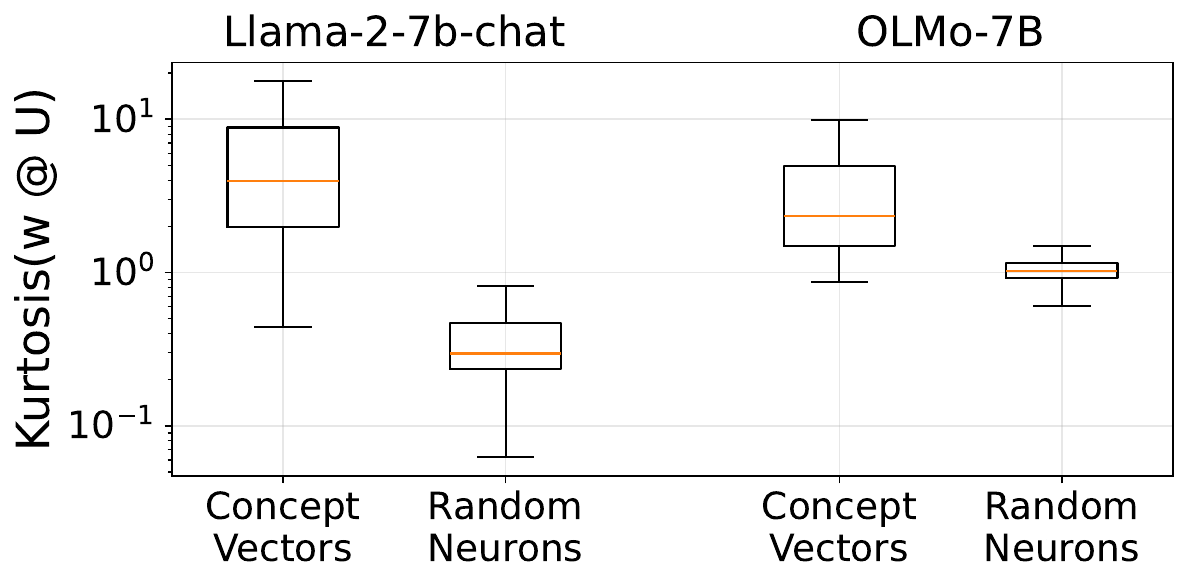}
    \caption{Vocabulary kurtosis of concept vectors in $\mathbf{W}_{\text{out}}$ \citep{conceptvec} vs.\ random neurons from the same layers.}
    \label{fig:conceptkurt}
\end{wrapfigure}
\paragraph{High kurtosis as a monosemanticity signal}
Given the above observations, we hypothesize that the distribution over the vocabulary induced by a weight vector could indicate how monosemantic it is. 
Specifically, we expect that monosemantic neurons will be correlated with higher kurtosis values of their vocabulary projections.
To test this, we compare the vocabulary kurtosis values of the concept vectors found by \citet{conceptvec} with those of randomly sampled neurons of the same layers. Figure~\ref{fig:conceptkurt} shows that, for both Llama-2-7B and OLMo-7B, vocabulary kurtosis creates a clear separation between these groups of neurons.
The median concept vector lies at the 90th percentile for Llama-2-7B and the 95th percentile for OLMo-7B relative to the randomly sampled neurons.
As further validation of vocabulary kurtosis being a meaningful signal, we tracked its values during pre-training in OLMo-2-1124-7B \citep{olmo2}. Our analysis shows that vocabulary kurtosis rises sharply in early training and concentrates in middle and final layers — confirming it is a learned property rather than an artifact (see \S\ref{appx:kurtosis_signal} for details).
Together, these observations motivate our approach: \textbf{low-kurtosis (polysemantic) neurons may be composed of multiple high-kurtosis (monosemantic) directions, 
which could be disentangled by maximizing non-Gaussianity}.

\section{ROTATE}
\label{sec:method}

We now introduce \methodname{}, a data-free method that, given a neuron weight vector $\mathbf{w}$, learns a set of rotation matrices $\{\mathbf{R_i}\}$, each yielding a \textit{vocabulary channel} $\mathbf{v}_i := \mathbf{R_i}\mathbf{w} $ that describes a monosemantic direction of $\mathbf{w}$.
An algorithm describing the method is provided in \S\ref{appx:rotate_alg}.

\paragraph{Optimization objective}
The core of our approach is in finding a rotation matrix $\mathbf{R} \in \mathbb{R}^{d \times d}$ such that the rotated vector $\mathbf{v} = \mathbf{w}\mathbf{R}$ 
will exhibit a high-kurtosis logit distribution $\mathbf{z} = \mathbf{v}\mathbf{U}$.
To steer the optimization towards interpretable features while maintaining fidelity to the neuron, we minimize a loss function $\mathcal{L}$ composed of two competing terms:
(a) \textbf{kurtosis loss} ($\mathcal{L}_{\text{kurt}}$), maximizing the kurtosis of $\mathbf{z}$ to push $\mathbf{w}$ towards monosemantic directions, and
(b) \textbf{regularization loss} ($\mathcal{L}_{\text{reg}}$), penalizing the cosine distance between $\mathbf{v}$ and $\mathbf{w}$.
This regularization anchors the discovered channels in $\mathbf{w}$, preventing convergence to arbitrary high-kurtosis directions.
\begin{equation}
\label{eq:loss}
    \mathcal{L} = -\lambda \cdot \mathcal{L}_{\text{kurt}} + \mathcal{L}_{\text{reg}} = -\lambda \cdot \log\!\left(1 + \text{Kurt}({\mathbf{z}})\right) + 1 - \frac{\mathbf{w} \cdot \mathbf{v}}{\|\mathbf{w}\| \|\mathbf{v}\|}
\end{equation}

We minimize $\mathcal{L}$ via gradient descent over a Householder parameterization of $\mathbf{R}$ \citep{householder}, which
enforces orthogonality by construction.
Let $\mathbf{h} \in \mathbb{R}^d$ be a learned vector, initialized as $\mathbf{h} \sim \mathcal{N}(0, I)$,
we define $\mathbf{R}$ as:
\begin{equation}
    \mathbf{R} = \mathbf{I} - 2\frac{\mathbf{h}\mathbf{h}^T}{\|\mathbf{h}\|^2}
\end{equation}
This parameterization allows us to optimize a $d$-dimensional vector that creates a full rank reflection matrix. Notably, a single Householder matrix is technically a reflection, yet we find it sufficient (see details in \S\ref{appx:ablations} and \S\ref{appx:compute} for method efficiency).

\paragraph{Iterative algorithm}
Optimizing Eq.~\ref{eq:loss} yields a single vocabulary channel. 
Since neurons often capture multiple concepts \citep{monosemanticity, poly-capacity, haystack}, we apply the optimization iteratively. However, naively repeating independent runs converges to the same local optimum (\S\ref{appx:convergence_analysis}), so we employ an iterative masking procedure.\footnote{We also investigated other strategies but found token masking to be most consistent (see \S\ref{appx:ablations}).}
After each iteration, we identify the tokens contributing most significantly to the channel's kurtosis and mask them to prevent re-discovery.
Let $\mathbf{z} = \mathbf{v}\mathbf{U}$ be the logit vector of the discovered channel with mean $\mu_{\mathbf{z}}$ and standard deviation $\sigma_{\mathbf{z}}$.
We mask high-contributing tokens with logit magnitudes exceeding $k$ standard deviations:
\begin{equation}
    \mathcal{T} = \{i : |z_i - \mu_{\mathbf{z}}| > k \cdot \sigma_{\mathbf{z}}\},
\end{equation}

This forces subsequent iterations to discover new high-kurtosis directions. We also mask known ``glitch tokens'' \citep{glitch,fishing}, which are under-trained embeddings whose extreme norms act as degenerate attractors (see \S\ref{appx:glitch}).
Each rotation $\mathbf{R}_i$ is optimized until loss convergence or a maximum step count.

\section{Experiments}{\label{sec:experiments}}

A natural question that arises is whether the weight-derived directions by \methodname{} capture the neuron's behavior during inference. 
To tackle this, we conduct evaluations along two axes: \textit{faithfulness}, i.e., how accurately the discovered channels predict the neuron's 
activation patterns (input-side) and concept promotion (output-side), and \textit{completeness}, i.e., how well the discovered channels explain the neuron's activation spectrum.
We find that \methodname{}'s data-free channels obtain consistently higher faithfulness and completeness scores than data-driven SAE baselines, explaining a larger fraction of the neuron's behavior. Moreover, channel ablations causally affect the neuron's activations on specific examples, while preserving its activations on other examples.
Additional evaluations of \methodname{} show that it finds the same vocabulary channels across different initializations (see \S\ref{appx:consistency}).

\subsection{Experimental setup}
\label{subsec:exp_setup}

The weight vectors $\mathbf{w}_{\text{gate}}$ and $\mathbf{w}_{\text{in}}$ of a neuron can be viewed as ``readers'' from the residual stream and $\mathbf{w}_{\text{out}}$ as the ``writer'' \citep{kv-memory}.
In our experiments, we apply \methodname{} to $\mathbf{w}_{\text{gate}}$ for the input side and $\mathbf{w}_{\text{out}}$ for the output side, running $n_{\text{iter}} = 50$ iterations per weight vector which achieves high reconstruction (cosine similarity $>0.95$, relative norm $>0.7$), see \S\ref{appx:reconstruction} for analysis).
We focus on $\mathbf{w}_{\text{gate}}$ rather than $\mathbf{w}_{\text{in}}$ for the input side as the gating activation is mostly positive, which simplifies the analysis, but \methodname{} is equally applicable to $\mathbf{w}_{\text{in}}$.
Hyperparameters are selected via grid search on a disjoint set of neurons (see \S\ref{appx:hyperparams} for details).
Using this configuration, we apply \methodname{} to Gemma-2-2B-it \citep{gemma2} and Llama-3.1-8B-Instruct \citep{llama3}. 
We focus on two layers per model, selected for their alignment with the vocabulary basis:\footnote{Evaluations of \methodname{} across early, middle, and late layers in both models show that it consistently scores above the random baseline (\S\ref{appx:layer_sweep}), with the strongest gains concentrated in middle-to-late layers, consistent with the training dynamics of vocabulary kurtosis (\S\ref{appx:kurtosis_signal}).} in Gemma, which uses tied embeddings (i.e., $E=U^T$), we analyze an early and a middle layer (layers 4 and 18) where weight vocabulary projection is geometrically valid; in Llama, we evaluate middle-to-late layers (layers 18 and 22), where the residual stream is most aligned with the unembedding matrix \citep{logitlens,kv-memory,shared}.  
From each layer we sample 100 random neurons. Examples of obtained channels are provided in \S\ref{appx:qualex}.

Let $\mathcal{C} = \{\mathbf{v}_1, \ldots, \mathbf{v}_k\}$ be the set of channels obtained for a neuron, given an input residual stream vector $\mathbf{x}$, we define the \emph{top channel} as $\mathbf{v}^* := \arg\max_{\mathbf{v} \in \mathcal{C}} (\mathbf{x} \cdot \mathbf{v})$, i.e., the channel most aligned with $\mathbf{x}$.

\paragraph{Evaluation data}
To validate the behavior of the extracted channels during inference on inputs, we collect a dataset $\mathcal{D}$ of 2 million tokens from the Pile \citep{dapile}, recording each token's residual stream vector before the MLP layer and the corresponding neuron activations. This dataset is used in our experiments
for retrieving top-activating examples and computing 
channel--example alignments.

\paragraph{Channel descriptions}
To evaluate channels, we first produce a natural-language description
for each one. Following 
\citet{output-enhancing}, we prompt an LLM with two sources of
evidence: the top-50 tokens in the channel's vocabulary projection
and its top activating examples from $\mathcal{D}$
(see \S\ref{appx:channel-desc} for the full prompt).

\subsection{Input-side channel faithfulness}\label{subsec:input-side}

Following automated interpretability protocols \citep{bills, transluce, paulo-auto}, we test whether the concept captured by a channel activates its corresponding neuron. 
Adopting the evaluation setup of \citet{assessing}, 
given a channel description, we prompt an LLM to create two sets of examples: \textit{activating} examples that match the description and \textit{neutral} examples that do not.
We then pass both sets through the model and record each neuron's maximum activation across token positions per example. 
This yields two sets of activation values per neuron $A_{\text{activating}}$ and $A_{\text{neutral}}$. A channel is considered faithful if $\mathbb{E}[{a \in A_{\text{activating}}}] > \mathbb{E}[{a \in A_{\text{neutral}}}]$, evaluated via a one-sided t-test ($p<0.05$) with 40 samples in each set.
Namely, the channel captures a concept that activates the neuron more strongly than other concepts.

As existing interpretability methods do not disentangle individual neuron weights into fine-grained components,
we adapt Gemma Scope and Llama Scope SAEs \citep{gemmascope, llamascope} trained on residual stream activations.
Concretely, we follow the weight-disentangling procedure of \citet{gur-arieh-etal-2025-precise}, which decomposes neuron weights by selecting the top-$k$ features of a trained residual-stream SAE most aligned with $\mathbf{w}$;
this is done by computing the dot product between $\mathbf{w}$ and each feature vector in the SAE's encoder and selecting the top-$k$ features with the highest alignment (see \S\ref{appx:extra_baselines} for more details).
These features serve as counterparts to \methodname{}'s vocabulary channels. 
We describe the selected features with two approaches, with their difference isolating the effect of the channel/feature discovery method from the description generation procedure:
\begin{itemize}
[leftmargin=*,topsep=0pt,itemsep=3pt,parsep=0pt]
    \item \textbf{SAE-Neuronpedia}: Descriptions from Neuronpedia \citep{neuronpedia} produced by prompting GPT-4 \citep{gpt4} with each feature's top-activating examples.
    \item \textbf{SAE-TopK}: Descriptions generated using the same procedure applied to \methodname{} channels (\S\ref{subsec:exp_setup}), collecting the top tokens from the feature's vocabulary projection and the top activating examples, then prompting an LLM to produce a description.
\end{itemize}

\begin{table}[t]
\setlength{\belowcaptionskip}{-15pt}
    \centering
    \footnotesize
    \begin{tabular}{l cccc cccc}
        & \multicolumn{4}{c}{\textbf{Faithfulness}} 
        & \multicolumn{4}{c}{\textbf{Completeness}} \\
        & \multicolumn{2}{c}{Llama-3.1} & \multicolumn{2}{c}{Gemma-2}
        & \multicolumn{2}{c}{Llama-3.1} & \multicolumn{2}{c}{Gemma-2} \\
        \cmidrule(lr){2-3} \cmidrule(lr){4-5} \cmidrule(lr){6-7} \cmidrule(lr){8-9}
        \textbf{Method} 
        & $\ell=18$ & $\ell=22$ & $\ell=4$ & $\ell=18$
        & $\ell=18$ & $\ell=22$ & $\ell=4$ & $\ell=18$ \\
        \midrule
        \methodname{} (Ours) & \textbf{0.71} & \textbf{0.58} & \textbf{0.46} & \textbf{0.47} & \textbf{0.55} & \textbf{0.49} & \textbf{0.55} & \textbf{0.60} \\
        SAE-Neuronpedia       & 0.45 & 0.41 & 0.33 & 0.35 
                                      & 0.44 & 0.41 & 0.42 & 0.49 \\
        SAE-TopK              & 0.49 & 0.46 & 0.34 & 0.37 
                                      & 0.40 & 0.40 & 0.36 & 0.42 \\
        Random                & 0.25 & 0.20 & 0.17 & 0.24 
                                      & 0.20 & 0.20 & 0.20 & 0.20 \\
        \bottomrule
    \end{tabular}
    \caption{Average Faithfulness and Completeness scores. \methodname{} consistently outperforms SAE-based baselines across models and layers. Random reflects chance-level performance.}
    \label{tab:faithfulness_completeness}
\end{table}

Table~\ref{tab:faithfulness_completeness} presents the faithfulness scores, showing that \methodname{} consistently outperforms the SAE baselines (0.46--0.71 vs. 0.33--0.49). 
The advantage is most pronounced in layer 18 of Llama-3.1 (0.71 vs. 0.49), 
likely because middle layers develop the strongest vocabulary-aligned structure (see analysis in \S\ref{appx:kurtosis_signal}), providing a richer signal for \methodname{}'s kurtosis-based optimization. In contrast, the gap narrows in layer 4 of Gemma-2 (0.46 vs. 0.34), where early-layer neurons may encode more distributed representations that are harder to disentangle.
The gap between \methodname{} and SAE-based methods suggests that weight-derived channels describe neuron activations more accurately than residual stream features extracted from SAEs. Notably, all methods substantially exceed the random baseline, confirming that both approaches capture meaningful structure, though \methodname{} captures it more precisely.

In addition to these results, we also evaluated two weight-space baselines on Gemma-2: \textbf{TopK-Iter}, an iterative logit-lens decomposition that repeatedly takes and masks the top tokens of $\mathbf{w}\mathbf{U}$, and \textbf{FastICA} \citep{fastica} that shares \methodname{}'s kurtosis-deflation structure. Both baselines performed at or below chance; see \S\ref{appx:extra_baselines} for details and results.

\paragraph{Causal validity via channel ablation}\label{subsec:input_ablation}
\begin{wrapfigure}{r}{0.5\columnwidth}
    \vspace{-12pt}
    \centering
    \includegraphics[width=0.44\columnwidth]{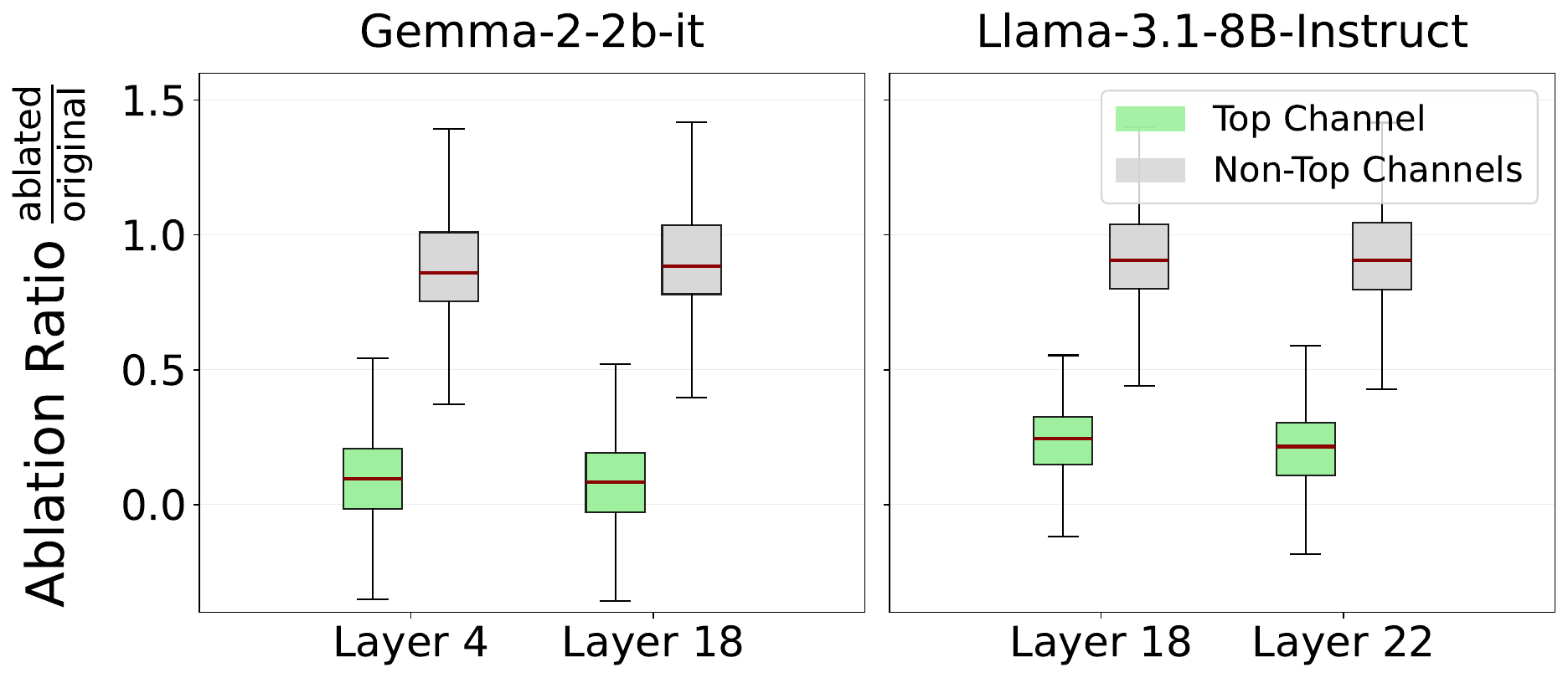}
    \caption{Input-side causal validity. Ablating the neuron's top channel drives its activation toward~0; ablating other channels leaves it near~1.}
    \label{fig:input_ablation}
    \vspace{-10pt}
\end{wrapfigure}
To test whether channels are causally responsible for the neuron's activation, we ablate the channel $\mathbf{v}$ from the neuron's weight vector $\mathbf{w}$ by projecting out its contribution: $\mathbf{w}_{\text{ablated}} = \mathbf{w} - (\mathbf{w} \cdot \mathbf{v})\,\mathbf{v}$.
Then, we compare the neuron activations before and after ablation.
Intuitively, if the channel controls a specific part of the neuron's behavior, then removing it should suppress activations on inputs related to that channel, while leaving other activations intact.

For each weight vector $\mathbf{w}$, we retrieve its top-1{,}000 activating examples from $\mathcal{D}$ and assign each example $\mathbf{x}$ to its top channel $\mathbf{v}^*$ (see \S\ref{subsec:exp_setup}).
Then, we ablate $\mathbf{v^*}$ from $\mathbf{w}$ and compute the ablation ratio, defined as the ratio between the ablated neuron's activation and the original activation for $\mathbf{x}$.
We measure this ratio on two sets of examples: those assigned to $\mathbf{v^*}$ and those assigned to other channels.

Figure~\ref{fig:input_ablation} shows that ablating the activated channel drives the ratio toward $0$ (\textcolor{mygreen}{\textbf{green}}), confirming that the channel is responsible for the neuron's firing on those inputs. Ablating a non-activated channel leaves the ratio near $1$ (\textcolor{mygray}{\textbf{gray}}), indicating that different channels do not interfere with one another. 
This shows that the discovered channels are both causally relevant and well-separated, with each governing a distinct subset of the neuron's behavior.

\subsection{Output-side channel faithfulness}
While input-side channels are selectively activated by different inputs, output-side channels all contribute simultaneously when the neuron fires. 
Thus, to evaluate faithfulness of output-side channels, we test what concepts the neuron promotes and whether ablating certain channels removes the expression of their concepts through the neuron.

We apply channel ablation as in \S\ref{subsec:input_ablation}, now targeting channels in $\mathbf{w}_{\text{out}}$.
To assess the effect of ablating a channel $\mathbf{v}$, we leverage the Patchscopes framework \citep{patchscopes} to decode information from $\mathbf{w}_{\text{out}}$ and the ablated vector $\mathbf{w}_\text{ablated}$.
Specifically, 

we feed 
to the model: 
$\texttt{"cat} \rightarrow \texttt{cat};\;
 \texttt{135} \rightarrow \texttt{135};\;
 \texttt{hello} \rightarrow \texttt{hello};\texttt{"}$ followed by either $\mathbf{w}_{\text{out}}$ or $\mathbf{w}_\text{ablated}$.
The few-shot format and conditioning the generation on the weight vector push the model to decode information from it.
Now, let $T_\mathbf{v}$ denote the set of top-$50$ tokens in the vocabulary projection of the channel $\mathbf{v}$.
We decode each of $\mathbf{w}_{\text{out}}$ and $\mathbf{w}_{\text{ablated}}$ multiple times, 
pooling all generated tokens per vector. Then, we compute the fraction of decoded tokens that belong to $T_\mathbf{v}$ in each pool, denoted $f_{\text{out}}$ and $f_{\text{ablated}}$, respectively, and report the relative change
$\Delta = (f_{\text{ablated}} - f_{\text{out}}) / f_{\text{out}}$.
For more details, see \S\ref{appx:patchscopes}.
We compare two ablations: \textit{self-channel ablation}, where we ablate the channel whose token set $T_\mathbf{v}$ we monitor, and \textit{cross-channel ablation}, where we ablate a different channel from the same neuron.
If the channels are causally disentangled, self-channel ablation should suppress the channel's tokens while cross-channel ablation should leave them intact.

\begin{wraptable}{r}{0.48\textwidth}
    \vspace{-1.2em}
    \footnotesize
    \centering
    \setlength{\tabcolsep}{3.5pt}
\begin{tabular}{@{}llcc@{}}
    \toprule
    \textbf{Model} & \textbf{Layer} & \textbf{Self (\%)} & \textbf{Cross (\%)} \\
    \midrule
    Gemma-2 & 4  & $-90 \pm 34$ & $+24 \pm 55$ \\
    -2b-it   & 18 & $-87 \pm 40$ & $+24 \pm 57$ \\
    \midrule
    Llama-3.1 & 18 & $-90 \pm 35$ & $+15 \pm 60$ \\
    -8B       & 22 & $-88 \pm 37$ & $+14 \pm 60$ \\
    \bottomrule
\end{tabular}
    \caption{Output-side causal validity via channel ablation. Mean ($\pm$ std) \% change in token frequency after self- or cross-channel ablation.}
    \label{tab:output_ablation}
    \vspace{-1em}
\end{wraptable}
Table~\ref{tab:output_ablation} 
presents the results. Self-channel ablation leads to near-complete suppression of the corresponding tokens (from $-87\%$ to $-90\%$). In contrast, cross-channel ablation slightly increases the frequency
(from $+14\%$ to $+24\%$), suggesting that a channel's tokens become more prominent when competing channels are removed.
This confirms that the discovered output channels are causally separated; each independently controls its corresponding concept, and removing one does not collapse the neuron's other functions.

\subsection{Decomposition completeness}\label{subsec:completeness}
The previous evaluations focused on whether a channel faithfully captures the behavior of its neuron. A question that remains is how many of the neuron behaviors do channels cover.
We approach this by evaluating \emph{completeness}, measuring how well the set of discovered channels collectively explains the neuron's activation landscape.
Specifically, we focus on input-side channels in $\mathbf{W}_{\text{gate}}$ which admit a natural test: given diverse inputs that activate the neuron, can we match each to an appropriate channel?\footnote{Output-side channels lack this structure; when a neuron activates, it promotes all its output channels, making it unclear how to attribute individual activations to specific channels.}

For every gate weight vector, we retrieve a sample of 100 out of its top-1000 activating input texts from $\mathcal{D}$ and, for each input $t$, identify its activated channel $\mathbf{v}^*$ (as defined in \S\ref{subsec:exp_setup}).
We then assess whether the description of $\mathbf{v}^*$ explains the neuron activation on $t$, for every such input-channel pair.
Using Gemini-3.1-Flash-Lite \citep{gemini3} as an LLM judge (see validation in \S\ref{appx:judge_validation}), we present the input text corresponding to $\mathbf{x}$ alongside five candidate channel descriptions: the description of $\mathbf{v}^*$ and four distracting descriptions sampled from channels of other neurons. The judge selects which description best explains why the neuron activated on this input.
We report \emph{matching accuracy}, defined as the fraction of examples where the judge selects the matched channel.
The full judge prompt and an example query are provided in \S\ref{appx:completeness}. 
We compare \methodname{} channels against random channels of other neurons, establishing a random baseline of 20\%, and the SAE-Neuropedia and SAE-TopK baselines from \S\ref{subsec:input-side}.

Table \ref{tab:faithfulness_completeness} presents the completeness scores. Across models and layers,
\methodname{} consistently outperforms the SAE baselines, achieving a matching accuracy of 49\%--60\% compared to 36\%--49\% for SAE features, both well above the 20\% chance level. 
For more than half of the neuron's top activating inputs, an LLM judge can correctly identify corresponding \methodname{} channel descriptions to the input, indicating that the discovered channels collectively cover the majority of the neuron's top activations.

\section{Enhancing neuron descriptions}\label{sec:enhancing}
In this section, we show that vocabulary channels can be leveraged to produce more comprehensive textual descriptions of neuron activations compared to existing pipelines.

\paragraph{Description generation}
\methodname{} produces dozens of channels per weight vector, raising the question of how to aggregate them into a single, coherent neuron description. 
Here, we experimented with four strategies, aggregating the descriptions of the first 25 channels from each of $\mathbf{w}_{\text{gate}}$ and  $\mathbf{w}_{\text{in}}$ 
(channel descriptions were obtained as in \S\ref{subsec:input-side}). 
From these strategies, we selected the following polarity-aware approach via a pairwise evaluation (see \S\ref{appx:description_generation} for details and results for all variants).
This approach exploits the distinct roles of the two weight vectors in the 
gated MLP: $\mathbf{w}_{\text{gate}}$ controls whether the neuron fires and 
$\mathbf{w}_{\text{in}}$ determines the activation's sign. We split $\mathbf{w}_{\text{in}}$ channels by their vocabulary projection skewness polarity and pair each group with all gate channels, yielding two per-neuron descriptions: one for positive and one for negative activations, each synthesized by Gemini-2.0-Flash (see \S\ref{appx:channel-synth}). 
Results below are from both polarities.

\paragraph{Baselines}
We compare \methodname{}-based descriptions against prominent baselines:
\begin{itemize}
[leftmargin=*,topsep=3pt,itemsep=3pt,parsep=0pt]
    \item \textbf{MaxAct+VocabProj}: We collect the neuron's 20 top-activating inputs from the Pile~\citep{dapile} and concatenate them with
    the top-50 vocabulary tokens in the projections of $\mathbf{w}_{\text{gate}}$ and $\mathbf{w}_{\text{in}}$.
    Then, we prompt Gemini-2.0-Flash to generate a concise description (see \S\ref{appx:description_generation} for the full prompt). This approach has been shown to outperform descriptions based on each source alone \citep{output-enhancing}.
  
    \item \textbf{MaxAct++}: As the strongest activation-based baseline, we use the descriptions by \citet{transluce} for neurons in Llama-3.1-8B-Instruct. These descriptions were generated via a multi-stage pipeline that involves the generation of candidate descriptions from top-activating inputs and scoring by a simulator that predicts per-token activations from a description.
    These automated descriptions have been shown to surpass human annotations on automated metrics.
\end{itemize}

\paragraph{Description evaluation}
We evaluate on 150 random neurons from Llama-3.1-8B-Instruct across 3 layers: 18 and 22 as in \S\ref{sec:experiments} and additionally layer 12 to test how the method performs in earlier layers.
To evaluate their descriptions in head-to-head comparisons we use Gemini-3-Flash \citep{gemini3} as a judge (see \S\ref{appx:judge_validation} for validation). 
Given an activating example and two candidate descriptions, the judge selects which description better explains the activation.
To control for position bias, we run each comparison twice with swapped order. We declare a winner when both orderings agree and otherwise a tie.
We evaluate descriptions on three setups:
(a) \textit{top 100 Pile activating inputs}, testing if descriptions capture the neuron's most pronounced behavior; (b) \textit{top 100-500 Pile activating inputs}, testing coverage beyond peak behavior; and (c) \textit{top 100 FineWeb activating inputs}, drawn from the MaxAct++ held-out test set \citep{fineweb}, testing generalization to a different data distribution. Pile evaluation examples are drawn from a disjoint subset not 
used for description generation.

\begin{figure}[t]
\setlength{\belowcaptionskip}{-10pt}
    \centering
    \includegraphics[width=\linewidth, height=0.18\textheight, keepaspectratio]{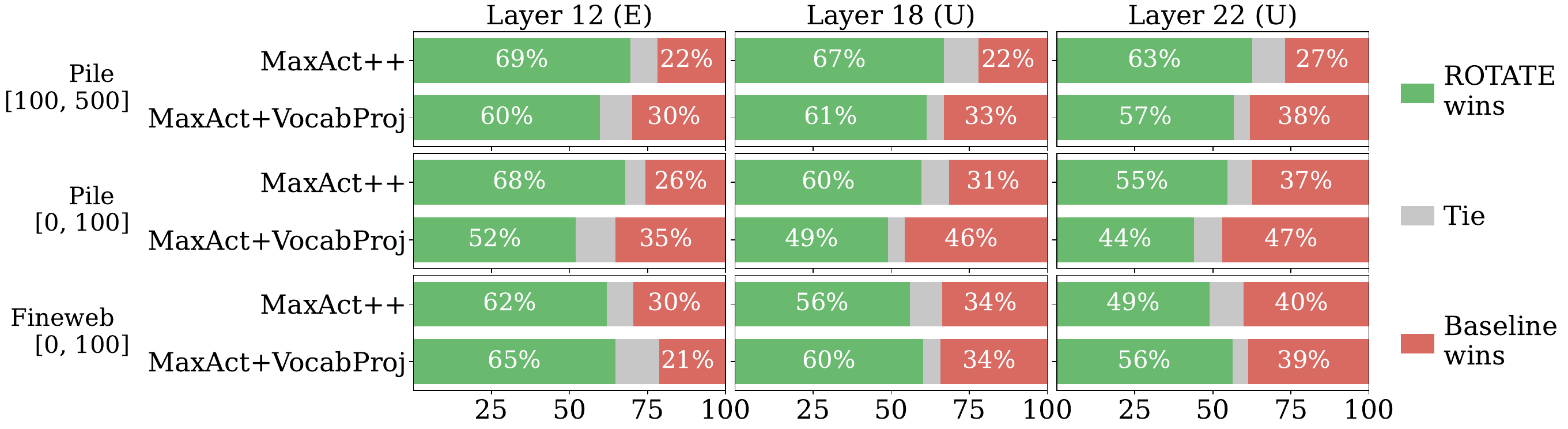}
\caption{Head-to-head pairwise evaluation of \methodname{} vocabulary channel descriptions 
against MaxAct+VocabProj and MaxAct++ baselines on Llama-3.1-8B-Instruct. Each bar shows the fraction of comparisons won by \textcolor{mywingreen}{\textbf{\methodname{}}}, \textcolor{mywingray}{\textbf{tied}}, or won by the \textcolor{mywinred}{\textbf{baseline}}.  Columns correspond to layers; rows to evaluation data sources and activation-rank 
ranges.}
    \label{fig:h2h}
\end{figure}

\paragraph{Results}
Figure~\ref{fig:h2h} shows the results, and examples are given in \S\ref{appx:wins-losses}.
\methodname{} wins against both baselines across nearly all setups. 
Against MaxAct++ the largest margins appear on moderate Pile activations (ranks 100--500), where \methodname{} achieves 63\%--69\% win rates, where MaxAct++ is furthest from its top-activation training regime.  
Against MaxAct+VocabProj, wins are most pronounced on the same moderate (ranks 100--500) range and on FineWeb (a different data distribution) while on top Pile activations the two methods are nearly tied. 
This reflects a basic trade-off: activation-based methods condition on extreme responses, giving strong signal for peak behavior but limited coverage elsewhere, whereas \methodname{} decomposes the weight vector independently of activation regime, naturally capturing concepts that surface at moderate levels.
These results demonstrate the practical gains of weight-derived vocabulary channels for neuron-level interpretability.

\section{Related work}\label{relwork}

Prior work has interpreted the weights of MLP layers \citep{kv-memory, ff-pred} and attention heads \citep{elhage2021mathematical, vocab-proj, maps} in the vocabulary space.
We build on this framework and learn rotations that disentangle neuron weights into monosemantic components.
Other works have identified underlying structures in MLP weights;
\citet{combinatorial} showed that MLPs in small networks can pack features via combinatorial ``feature channel codes'',
\citet{bilinear} found that bilinear MLPs can admit eigen-decomposition of their weights into interpretable components, and \citet{snmf} used MLP activations to discover neuron combinations that capture concepts and outperform SAEs on causal steering. 
Unlike these works, \methodname{} achieves \textit{data-free} decomposition of MLP layers in modern LMs.

Our study also relates to a large body of work on neurons in LMs \citep{neuron-survey}, and contributes to tackling the challenge of polysemanticity \citep{superposition,arora,haystack}.
While SAEs have been the dominant approach to recovering monosemantic units in LMs \citep{monosemanticity, saes, gao2025scaling}, they require large-scale activation data.
Recently, \citet{gur-arieh-etal-2025-precise} adapted residual-stream SAEs to decompose neuron weights. We compare against this approach and show that \methodname{} consistently outperforms it in faithfulness and completeness 
with respect to the neuron's behavior. \methodname{} also complements efforts to automatically describe neurons \citep{bills, transluce, maia, output-enhancing} by leveraging their fine-grained decompositions into channels.

Several methods discover interpretable structure by optimizing orthogonal transformations. DAS \citep{rotations-das} learns such matrices via supervised gradient descent to isolate causal features in the residual stream, and \citet{huang2026decomposing} learn them without supervision, partitioning activation space into interpretable subspaces by minimizing within-subspace nearest-neighbor distances over a large corpus of activations. \methodname{} learns similar rotations, but without supervision or data, operating entirely in weight space. 

Key components of \methodname{}---kurtosis maximization, an orthogonality constraint, and iterative deflation---mirror the structure of classical Independent Component Analysis \citep[ICA;][]{ica} and Projection Pursuit \citep{projectionpursuit}, in particular deflation-based algorithms such as FastICA \citep{fastica} and RobustICA \citep{robustica} that extract non-Gaussian components one at a time. The key difference lies in what the statistic is computed over. Classical ICA estimates an unmixing matrix from a data matrix of many observations, measuring non-Gaussianity across samples, whereas \methodname{} has no observations: it operates on a single static weight vector and measures kurtosis across the $V$ vocabulary logits of its projection through the fixed unembedding $\mathbf{U}$, removing exactly the data dependence that both DAS and ICA require.

\section{Conclusion and discussion}
We introduce \methodname{}, a data-free method that disentangles MLP neuron weights into interpretable vocabulary channels by maximizing kurtosis in the model's vocabulary space. The discovered channels provide faithful, causally meaningful descriptions of neuron behavior, outperforming SAE-based baselines in terms of faithfulness and completeness. Moreover, aggregating channel descriptions yields comprehensive neuron descriptions that achieve higher win rates over existing approaches. Taken together, vocabulary channels are positioned as a scalable, fine-grained unit of analysis for interpreting LMs.
Future work could leverage \methodname{} for more accurate, fine-grained circuit discovery and for studying interactions between network components. Further discussion on limitations is in \S\ref{appx:limitations}.

\section*{Acknowledgments}
We thank Ori Yoran and Guy Dar for valuable feedback, and Or Shafran, Clara Suslik, Daniela Gottesman, and Shir Rashkovits for their help with the evaluation of the LLM judge. This research was supported in part by the Academic Research Program at Google, Len Blavatnik and the Blavatnik Family foundation, the Alon Scholarship, a Coefficient Giving grant, and the Israel Science Foundation grant 1083/24.

\bibliography{rotating_neurons}
\bibliographystyle{colm2026_conference}

\appendix

\section{Additional preliminaries}\label{appx:add-pre}

\subsection{Kurtosis and Skewness}\label{appx:kurtosis-def}

Kurtosis is the fourth standardized moment of a distribution:
\begin{equation}
    \text{Kurt}(X) = \mathbb{E}\left[\left(\frac{X - \mu}{\sigma}\right)^4\right] - 3
\end{equation}
where $\mu$ and $\sigma$ are the mean and standard deviation of $X$. We subtract 3 so that a
Gaussian distribution has kurtosis zero (\emph{excess kurtosis}). Positive values indicate heavier
tails and a sharper peak than a Gaussian, meaning more of the variance is due to rare, extreme
values.

Skewness is the third standardized moment, measuring the asymmetry of a distribution:
\begin{equation}
    \text{Skew}(X) = \mathbb{E}\left[\left(\frac{X - \mu}{\sigma}\right)^3\right]
\end{equation}
Positive skewness indicates a heavier right tail (extreme positive logits dominate), while
negative skewness indicates a heavier left tail (extreme negative logits dominate). In our
setting, we use skewness polarity to distinguish channels that \emph{promote} tokens (positive
skewness) from those that \emph{suppress} them (negative skewness).

\begin{figure}[ht]
    \centering
    \includegraphics[width=0.95\linewidth]{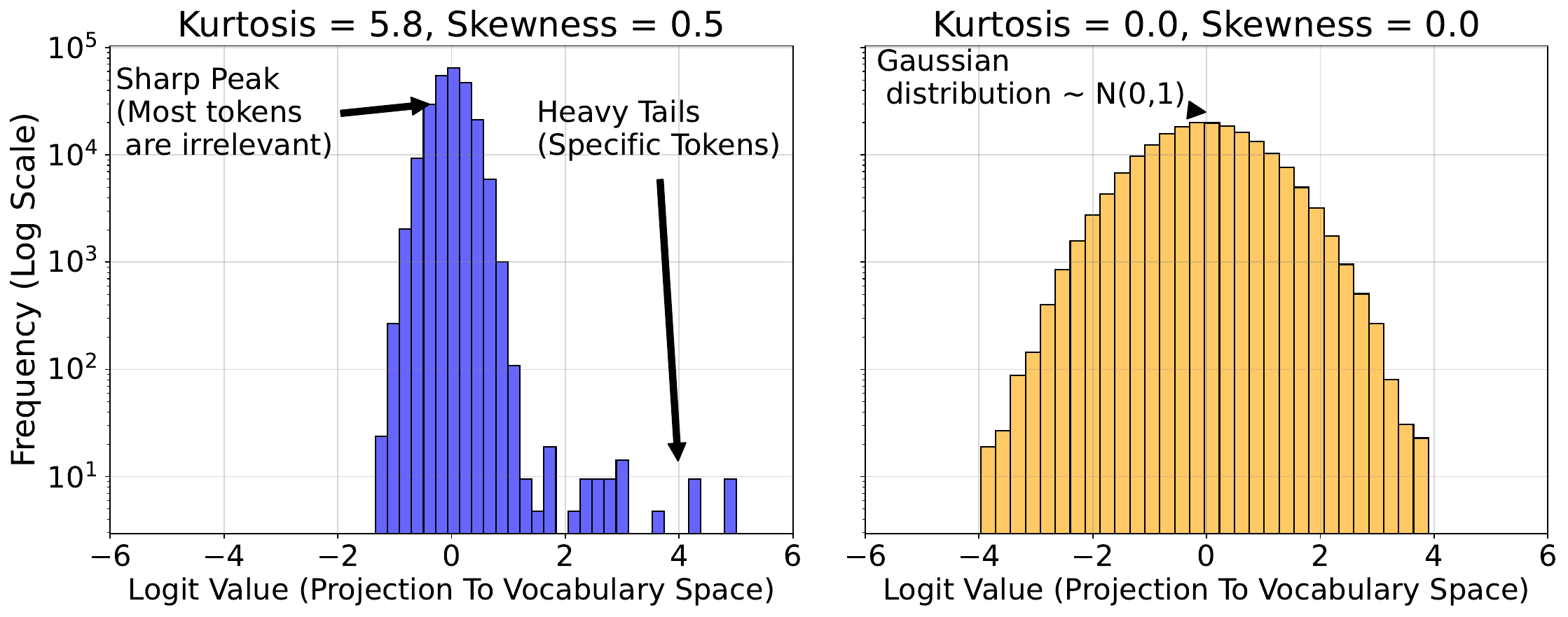}
    \caption{A distribution with high kurtosis and positive skewness, concentrated around zero with few extreme outliers (left), compared to a Gaussian (right).}
    \label{appx:kurtosis-exp}
\end{figure}

In our setting, we treat the logit vector $\mathbf{z} = \mathbf{w}\mathbf{U} \in \mathbb{R}^V$
as a distribution over the vocabulary: high kurtosis indicates that the neuron acts strongly on
a sparse set of tokens while having negligible effect on the rest, and the skewness sign
determines whether those tokens are promoted or suppressed.
Figure~\ref{appx:kurtosis-exp} illustrates this contrast.

\section{Vocabulary kurtosis across training and model families}\label{appx:kurtosis_signal}

\begin{figure}[ht]
    \centering
    \includegraphics[width=0.6\linewidth]{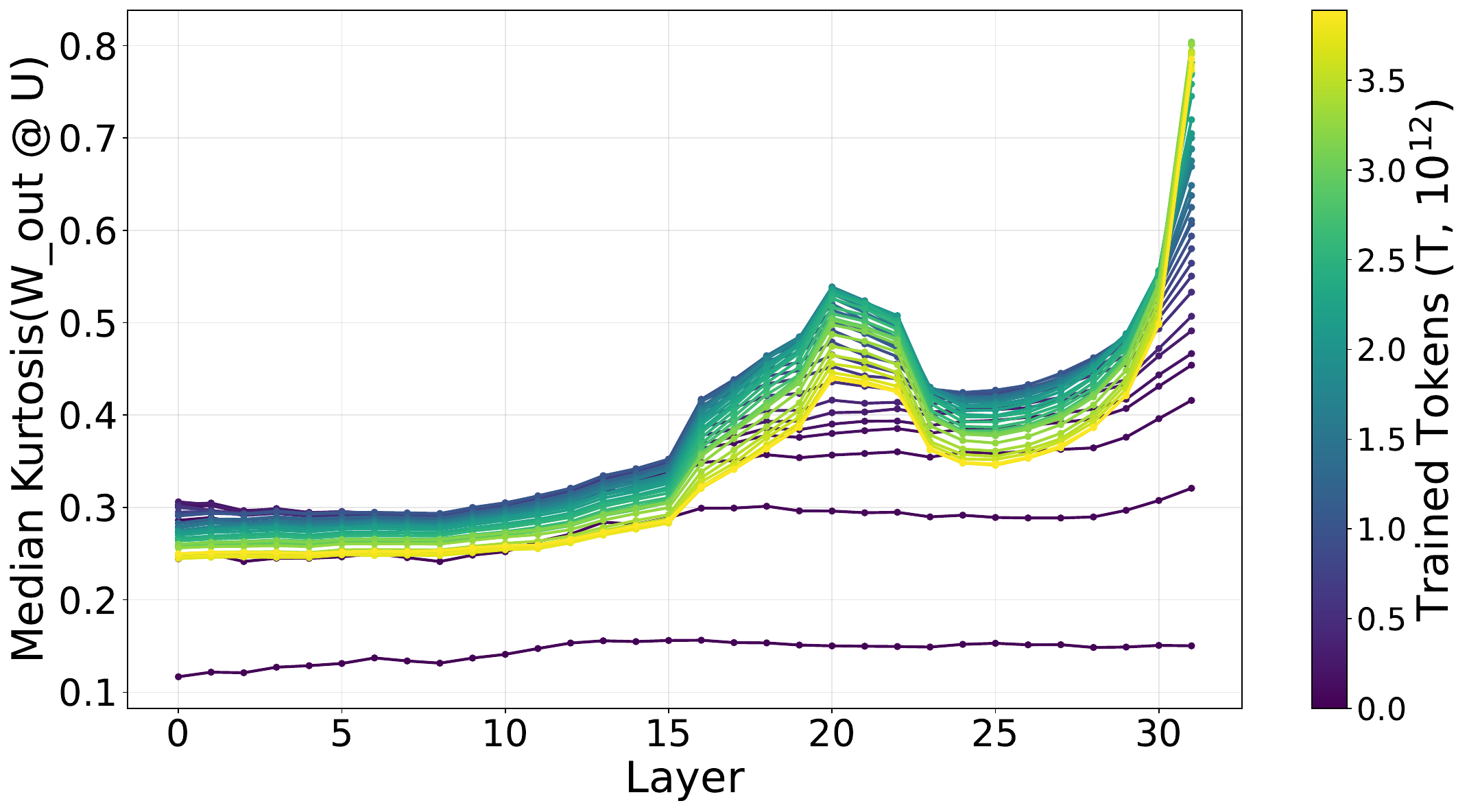}
    \caption{Median vocabulary kurtosis values of neuron weights in   $\mathbf{W}_{\text{out}}$ across layers and checkpoints of OLMo-2-1124-7B \citep{olmo2}. 
    We observe clear learning dynamics, rising sharply in early training and concentrating in middle and late layers. This temporal pattern confirms that vocabulary-aligned monosemantic structure is a learned property. 
    }
    \label{fig:kurt_evolution}
\end{figure}

\paragraph{Across training}
To verify that vocabulary kurtosis reflects genuinely learned structure rather than a static property of random initialization, we track its evolution during pre-training. Figure~\ref{fig:kurt_evolution} shows the median vocabulary kurtosis of $\mathbf{W}_{\text{out}}$ neurons in OLMo-2-1124-7B \citep{olmo2} across 4 trillion training tokens. At initialization, kurtosis values are near zero (consistent with Gaussian-distributed weights). During early training, median kurtosis rises sharply before stabilizing, with the strongest concentration emerging in middle layers (around layers 15--20) and the final layers. This temporal and layer-wise pattern confirms that vocabulary-aligned monosemantic structure is actively shaped by training.

\paragraph{Across model families}

\begin{figure}[ht]
    \centering
    \includegraphics[width=0.95\linewidth]{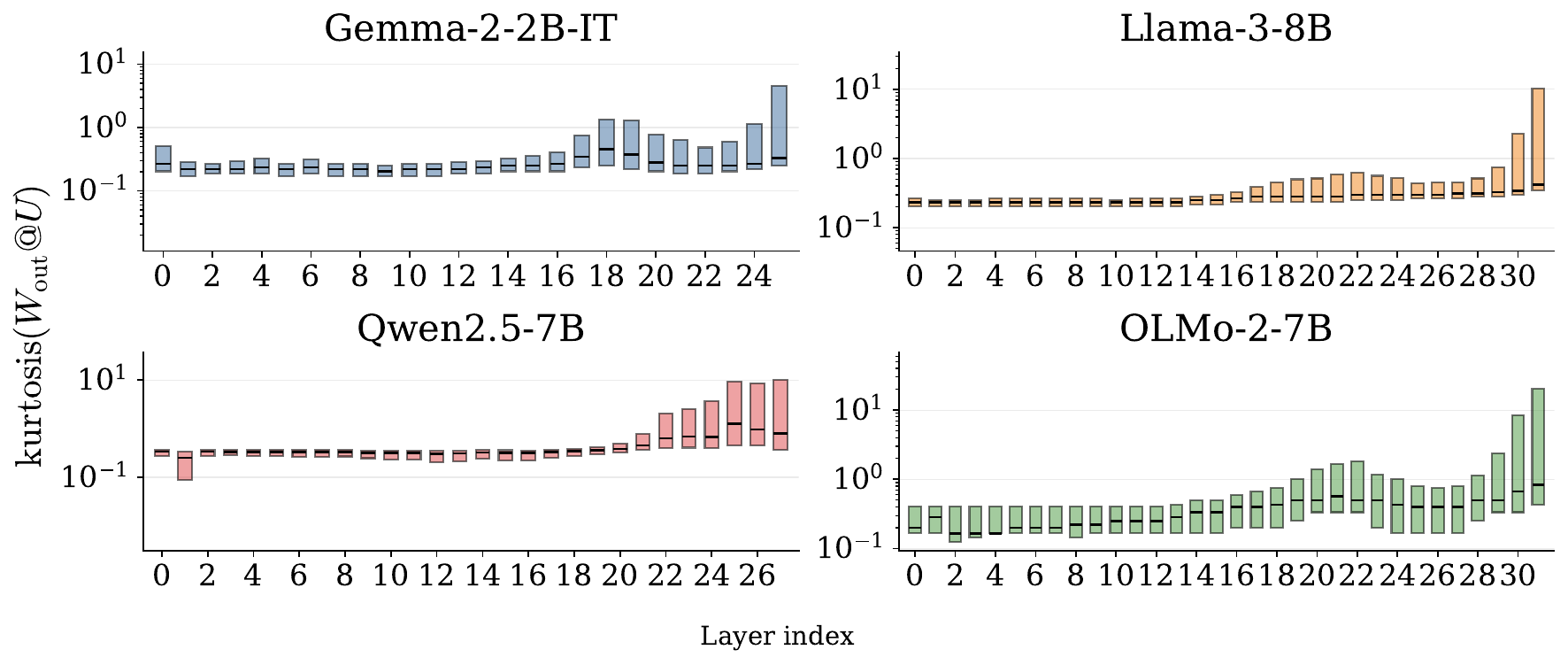}
    \caption{Per-layer vocabulary kurtosis distributions of $\mathbf{W}_{\text{out}}$ neurons for one representative model per family.}
    \label{fig:kurt_family}
\end{figure}

This layer-wise pattern, where middle-late and output-facing layers develop the strongest vocabulary-aligned structure, is consistent across multiple model families, as can be seen in Figure~\ref{fig:kurt_family}.

\section{\methodname{} additional details and results}\label{appx:rotate_alg}

\subsection{Algorithm}\label{appx:algorithm}
Algorithm~\ref{alg:kurtosis_rotation} provides the full pseudo-code for \methodname{}. Given a neuron weight vector $\mathbf{w}$ and the unembedding matrix $\mathbf{U}$, the method iteratively discovers vocabulary channels by optimizing Householder reflections to maximize vocabulary-space kurtosis. Each iteration yields a single channel; after discovery, the tokens driving its kurtosis are masked to force subsequent iterations toward new directions. The process terminates 
after $n_{\text{iter}}$ iterations. 
Below we provide additional details on implementation choices and design decisions.

\begin{algorithm}[ht]
\small
\caption{\methodname{}}
\label{alg:kurtosis_rotation}
\begin{algorithmic}[1]
\setlength{\baselineskip}{1.25em}
\STATE \textbf{Input:} MLP weight vector $\mathbf{w}$, unembedding matrix $\mathbf{U}$, kurtosis function $\gamma(x)$, kurtosis threshold $\tau$, learning rate $\eta$, $\lambda$, standard deviation magnitude $k$, $n_{\text{iter}}$, $n_{\text{step}}$.
\STATE \textbf{Output:} Set of discovered rotation matrices $\mathcal{R}$.

\STATE $\mathbf{m} \gets$ \texttt{init\_mask($\mathbf{U}$)}
\STATE $\mathcal{R} \gets \{\},\;\; i \gets 0 $ 

\REPEAT
    \STATE $i \gets i + 1$
    \STATE $\mathbf{h} \sim \mathcal{N}(0, I)$ \COMMENT{Random initialization}
    \STATE $\mathbf{R} \gets I - 2\frac{\mathbf{h}\mathbf{h}^T}{\|\mathbf{h}\|^2}$ \COMMENT{Householder reflection}
    \STATE \texttt{optimizer} $\gets$ \texttt{AdamW($\eta$)}
    \STATE $s \gets 0$

    \WHILE{$s < n_{\text{step}}$} 
        \STATE ${\mathbf{v}} \gets \mathbf{w}\mathbf{R} $ \COMMENT{Rotate $\mathbf{v}$ with $R$}
        \STATE $\mathbf{z} \gets {\mathbf{v}} \mathrm{U}$ \COMMENT{Obtain logits vector}
        \STATE $\hat{\mathbf{z}} \gets \mathbf{z} \odot \mathbf{m}$ \COMMENT{Mask tokens}
        
        \STATE $\mathcal{L}_{\text{kurt}} \gets \log(1 + \gamma(\hat{\mathbf{z}}))$ \COMMENT{Kurtosis loss}
        \STATE $\mathcal{L}_{\text{reg}} \gets 1 - \frac{\mathbf{v} \cdot {\mathbf{w}}}{\|\mathbf{v}\| \|{\mathbf{w}}\|}$ \COMMENT{Regularization loss}
        \STATE $\mathcal{L} \gets -\lambda \cdot \mathcal{L}_{\text{kurt}} + \mathcal{L}_{\text{reg}}$
        \STATE \texttt{optimizer.step($\mathcal{L}$)}
        \STATE $s \gets s + 1$
    \ENDWHILE
    
    \STATE $\mathcal{R} \gets \mathcal{R} \cup \{\mathbf{R}\} $
    \STATE $\mathcal{T} \gets \{i : |z_i - \mu_{\mathbf{z}}| > k \cdot \sigma_{\mathbf{z}}\}$ \COMMENT{High-kurtosis tokens}
\STATE $m_i \gets 0 \;\; \forall i \in \mathcal{T}$ \COMMENT{Mask discovered tokens}    
\UNTIL{$\gamma(\hat{\mathbf{z}}) < \tau$ or $i > n_{\text{iter}}$}
\STATE \textbf{return} $\mathcal{R}$
\end{algorithmic}
\end{algorithm}

\subsection{Weight reconstruction analysis}\label{appx:reconstruction}

The iterative nature of \methodname{} raises two termination questions: (1)~when to stop optimizing a single rotation matrix, and (2)~how many iterations to run per neuron. For~(1), we follow standard practice and terminate when the loss change falls below a threshold $\epsilon$ or a maximum step count $n_{\text{step}}$ is reached. For~(2), rather than attempting to estimate the ``polysemanticity degree'' of each neuron, we set a fixed iteration budget $n_{\text{iter}} = 50$ and verify empirically that this suffices for high-fidelity reconstruction.

To assess how well the discovered channels collectively reconstruct the original weight vector, we track two metrics across iterations, evaluated on Gemma-2-2B-it. Given channels $\{v_1, \dots, v_t\}$ discovered after $t$ iterations, we define the residual $r_t = w - \sum_{i=1}^{t} (w \cdot v_i) v_i$ and report: (1)~per-channel cosine similarity between each newly discovered channel $v_t$ and $w$, and (2)~cumulative explained norm, defined as $1 - \|r_t\| / \|w\|$.

\begin{figure}[ht]
    \centering
    \includegraphics[width=\linewidth]{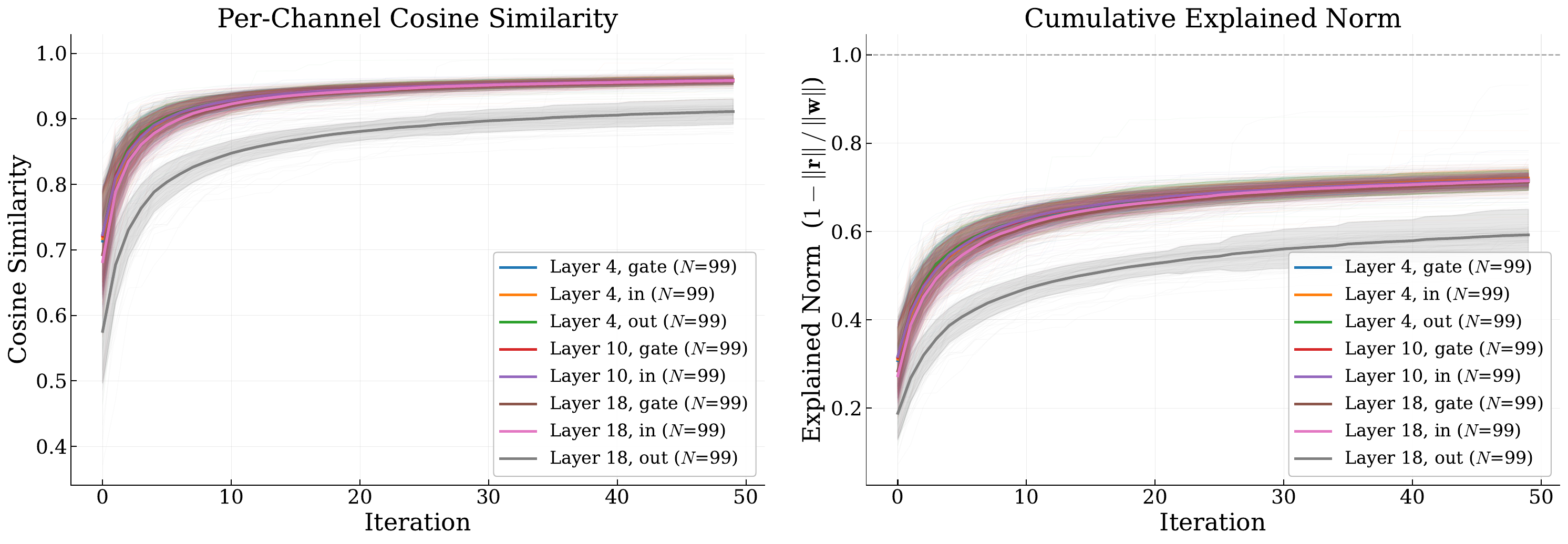}
    \caption{Weight reconstruction analysis on Gemma-2-2B-it. \textbf{Left:} Per-channel cosine similarity with the original weight vector $w$ across iterations. \textbf{Right:} Cumulative explained norm ($1 - \|r_t\| / \|w\|$) over iterations. Lines show medians across 99 neurons; shaded regions indicate inter-quartile ranges. Channels collectively reconstruct nearly all of $w$ within 50 iterations across all layers and weight types.}
    \label{fig:reconstruction}
\end{figure}

Figure~\ref{fig:reconstruction} shows both metrics for 99 randomly sampled neurons per layer and weight type. Early channels capture the dominant directions of $w$ (cosine similarity $> 0.9$ within ${\sim}10$ iterations), while later channels contribute smaller but consistent refinements. By iteration 50, the cumulative explained norm approaches 1.0 across all layers and weight types, confirming that 50 iterations suffice to account for nearly all of the original weight vector's norm. The consistent behavior across layers and weight matrices (gate, in, out) indicates that the decomposition is robust to the specific structure of the weight vector.

\subsection{Faithfulness and completeness across layers}\label{appx:layer_sweep}
To assess whether our main results are specific to the layers selected in \S\ref{subsec:exp_setup}, we run \methodname{} and the input-side evaluation suite on a sweep spanning early, middle, and late layers of both models.

Table~\ref{tab:layer_sweep} reports faithfulness and completeness per layer. Every evaluated layer remains above the random baselines (faithfulness 20--25\%, completeness 20\%), with the strongest gains concentrated in middle-to-late layers, mirroring the training dynamics of vocabulary kurtosis (Figure~\ref{fig:kurt_evolution}), which concentrates in the same layers.

\begin{table}[ht]
\centering
\small
\begin{tabular}{lcccc}
\toprule
 & \multicolumn{2}{c}{Gemma-2-2b-it} & \multicolumn{2}{c}{Llama-3.1-8B-Instruct} \\
\cmidrule(lr){2-3} \cmidrule(lr){4-5}
Layer & Faithfulness & Completeness & Faithfulness & Completeness \\
\midrule
2  & 46.9 & 40.7 & --   & --   \\
4  & 45.7 & 55.7 & 49.7 & 41.3 \\
8  & 45.2 & 43.5 & 49.3 & 35.8 \\
18 & 46.1 & 59.7 & 70.9 & 55.1 \\
22 & --   & --   & 57.5 & 49.3 \\
24 & 42.9 & 37.4 & --   & --   \\
\bottomrule
\end{tabular}
\caption{Layer sweep of input-side faithfulness and completeness (\%). All layers exceed the random baselines (faithfulness 20--25\%, completeness 20\%), with the strongest gains in middle-to-late layers.}
\label{tab:layer_sweep}
\end{table}

\subsection{Channel consistency}\label{appx:consistency} 

Since \methodname{} relies on a non-convex optimization procedure with random initialization (Algorithm~\ref{alg:kurtosis_rotation}),
we evaluate the stability of the algorithm’s output as an additional means of validating the method.

\paragraph{Experiment}
We run \methodname{} with 4 different random seeds on the same set of 50 randomly sampled gate neurons from layer 18 of Gemma-2-2B-it. For each neuron, this yields 4 independent sets of discovered channels. To quantify consistency, we measure whether the same channels are recovered across runs. For each pair of runs, we compute the pairwise cosine similarity between all channels from run A and all channels from run B. We then apply greedy matching to find the best one-to-one alignment between the two channel sets. For each matched pair, we compute the Jaccard similarity of their top-$k$ tokens to verify semantic agreement. High similarity across matched pairs indicates that the discovered vocabulary channels are stable features of the weight landscape.

\medskip
\noindent\begin{minipage}[t]{0.56\textwidth}
\paragraph{Results}
We report a mean cosine similarity of $0.9 \pm 0.04$ and a mean Jaccard similarity of $0.8 \pm 0.05$ across matched pairs. These high similarity scores demonstrate that \methodname{} consistently recovers the same semantic directions regardless of initialization. Figure~\ref{fig:consistent} shows an example for a pair of executions with the matching channels marked.
Notably, channels are not always discovered in the same order across runs, as they sometimes appear off-diagonal. This is expected as the random initialization of the Householder vector $\mathbf{h}$ determines which local optimum is found first, while the masking procedure ensures subsequent iterations discover different channels. The consistency of the set of discovered channels, despite varying discovery order, suggests these directions are genuine structures in the weight space rather than artifacts of a particular optimization trajectory.
\end{minipage}\hfill
\begin{minipage}[t]{0.40\textwidth}
    \centering
    \adjustbox{valign=t}{\includegraphics[width=\linewidth]{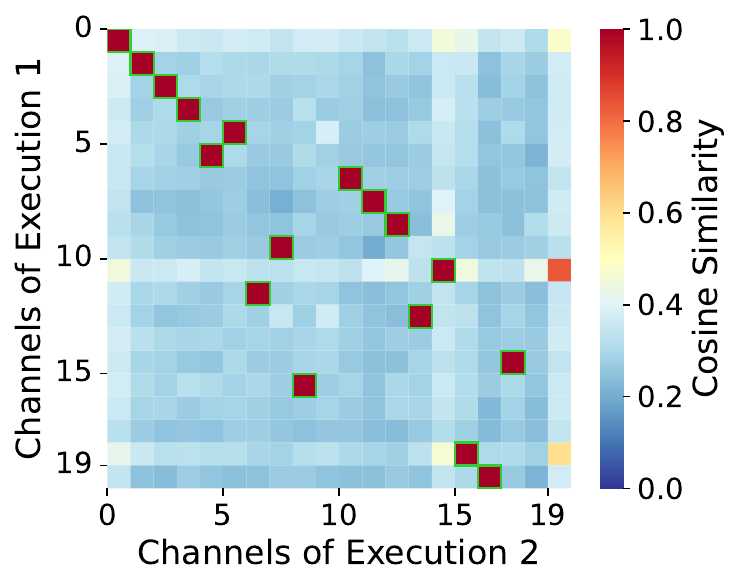}}
    \captionof{figure}{Consistency of \methodname{} across different initializations. The heatmap displays a pairwise cosine similarity between vocabulary channels discovered in two separate execution runs (Execution 1 vs.\ Execution 2) for the same target neuron.}
    \label{fig:consistent}
\end{minipage}

\subsection{Avoiding glitch tokens}\label{appx:glitch}
A practical challenge we encountered is that the optimization frequently converges to ``glitch tokens'' \citep{glitch}, which are under-trained token embeddings characterized by extreme norms.
Since our objective maximizes kurtosis, it is inherently sensitive to such outliers; the extreme norms of these tokens manifest as high-kurtosis directions that act as degenerate attractors in the optimization landscape. To prevent the algorithm from exploiting these tokenizer artifacts, we initialize the mask $\mathbf{m}$ (Alg.~\ref{alg:kurtosis_rotation}, line 3) to exclude known glitch tokens \citep{fishing} and ensure the method focuses on genuine semantic sparsity.

\subsection{Ablations}\label{appx:ablations}

\paragraph{Applying rotations on the same vector}
\label{appx:convergence_analysis}

To motivate the need for iterative token masking, we compare the standard \methodname{} pipeline with token masking between iterations against a variant that performs independent optimization runs with no depletion after each iteration. Meaning neither token masking nor residual subtraction between iterations.

We first demonstrate that without depletion, the optimization landscape contains a single dominant attractor. We run \methodname{} on 50 gate, in, and out neurons from Layer~18 of Gemma-2-2B-it, executing 20 independent optimization runs per neuron with different random seeds but no masking between runs. For each run, we record the anchor token (the top token of the vocabulary-projected channel) and the set of top-20 tokens. The mean pairwise Jaccard similarity of top-20 token sets is $0.60$, confirming strong semantic agreement even when the exact anchor token differs slightly.

\begin{figure}[t]
    \centering
    \includegraphics[width=0.98\linewidth]{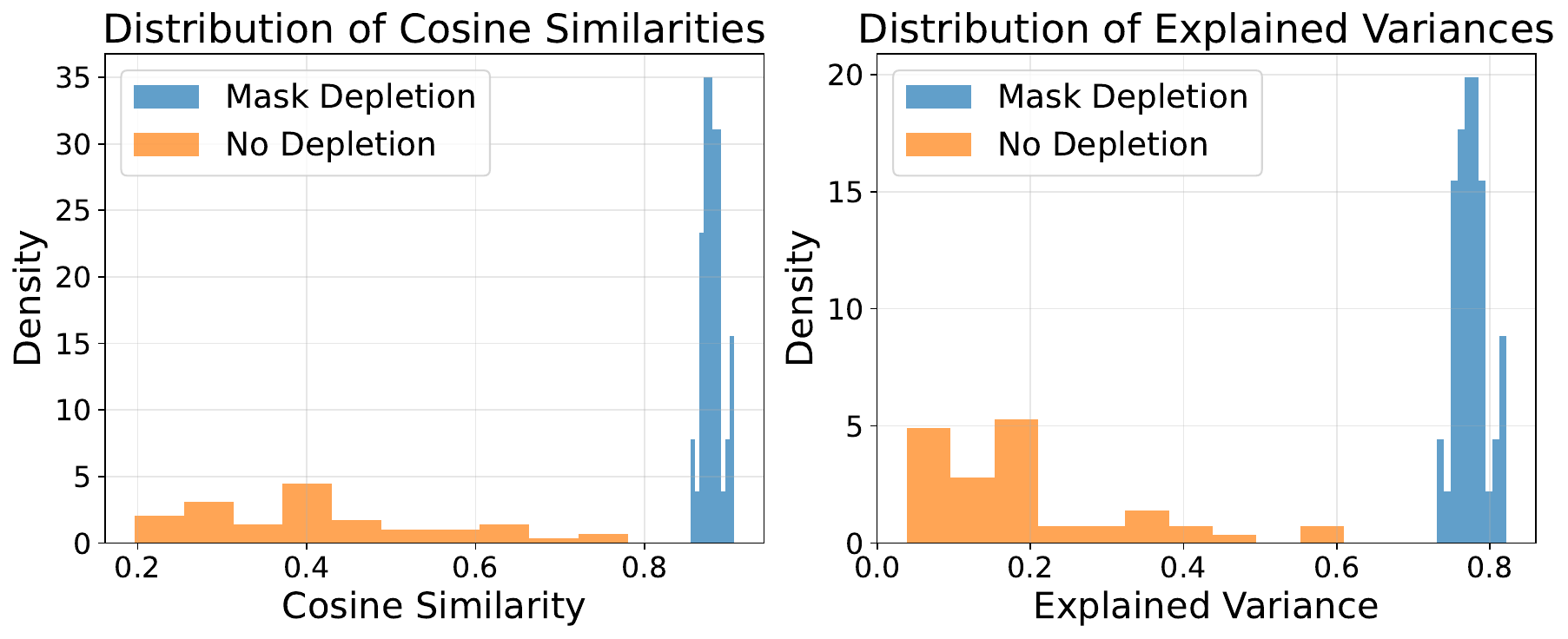}
    \caption{Effect of token masking on iterative decomposition quality. We compare \methodname{} with token masking against independent optimization runs with no depletion over 20 iterations on 50 gate neurons from Layer~18, Gemma-2-2B-it. \textbf{Left:} Per-channel cosine similarity with $\mathbf{w}$. \textbf{Right:} Cumulative explained norm ($1 - \|\mathbf{r}_t\|/\|\mathbf{w}\|$). Without masking, all iterations converge to the same dominant direction, yielding negligible reconstruction progress (mean explained norm $0.19$ vs.\ $0.78$).}
    \label{fig:masking_ablation}
\end{figure}

This redundancy directly harms decomposition quality. Figure~\ref{fig:masking_ablation} compares both variants over 20 iterations on the same set of gate neurons. Without depletion, nearly every iteration rediscovers the same dominant direction, yielding a mean cosine similarity of only $0.42$ and a mean explained norm of $0.19$, indicating that repeated runs contribute almost no additional reconstruction of $\mathbf{w}$. With token masking, subsequent iterations are steered toward novel high-kurtosis directions, achieving a mean cosine similarity of $0.88$ and a mean explained norm of $0.78$.
Consistent patterns hold for $\mathbf{w}_\text{in}$ and $\mathbf{w}_\text{out}$. These results confirm that depletion is essential: without it, the iterative procedure collapses to a single channel and fails to decompose the neuron.

\paragraph{Applying subtraction instead of masking}
\label{appx:ablation_subtraction}
To prevent the iterative optimization from rediscovering the same semantic directions, \methodname{} employs token masking. A standard alternative, common in methods like ICA, is iterative residual subtraction (deflation), where the projection of the discovered channel is subtracted directly from the weight vector before the next iteration. 

\begin{figure}[ht]
    \centering
    \includegraphics[width=\linewidth]{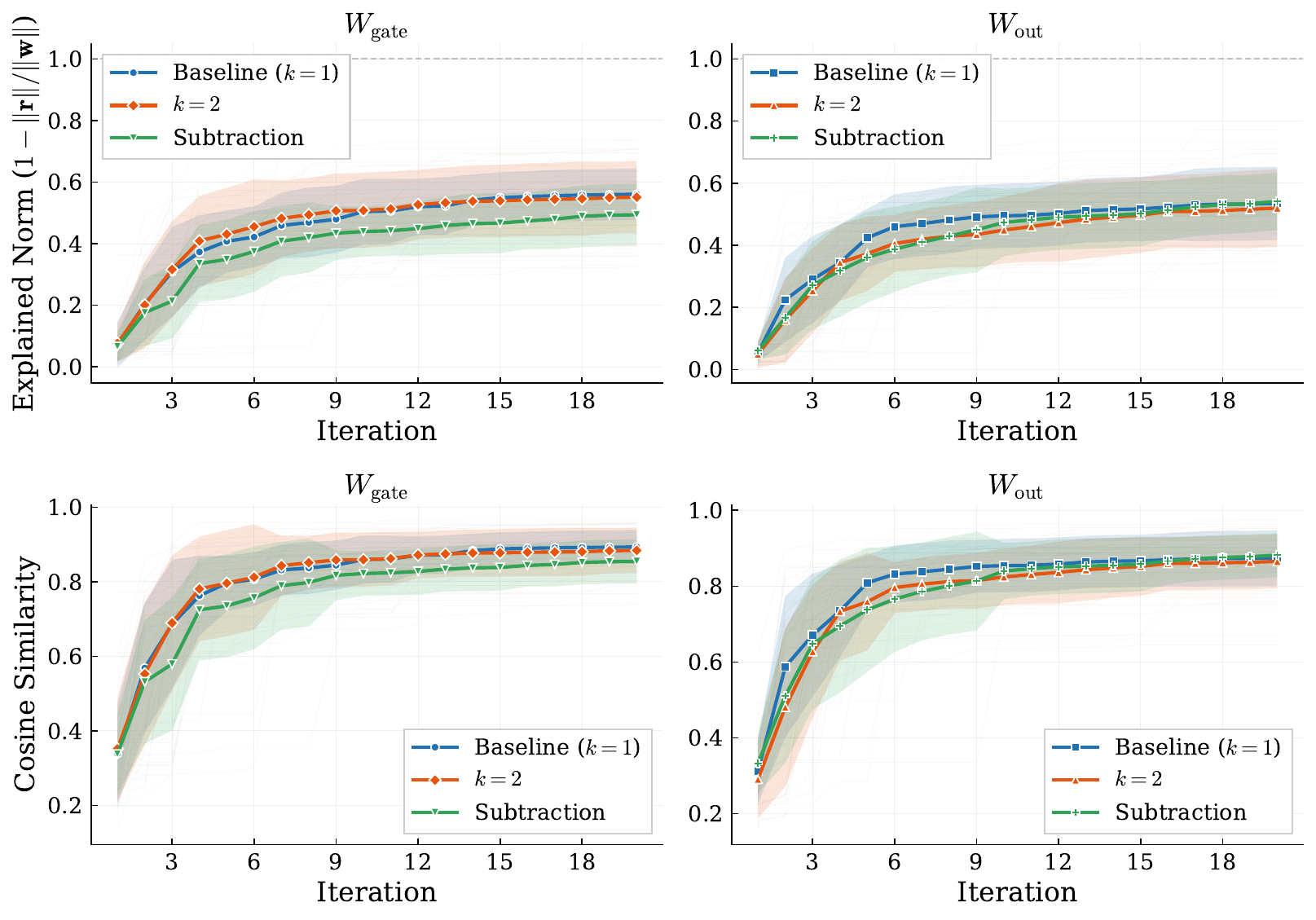} 
    \caption{Ablation results evaluating weight reconstruction across optimization iterations for $W_{\text{gate}}$ (left) and $W_{\text{out}}$ (right). We compare the \methodname{} baseline (token masking, $k=1$ Householder matrix) against two variants: utilizing a proper rotation via two Householder matrices ($k=2$), and using residual subtraction instead of token masking. \textbf{Top:} Cumulative Explained Norm ($1 - \|\mathbf{r}\| / \|\mathbf{w}\|$). \textbf{Bottom:} Cosine similarity between the reconstructed vector and the original weight vector. The baseline ($k=1$) matches the performance of the more complex $k=2$ parameterization and consistently outperforms residual subtraction.}
    \label{fig:ablations_reconstruction}
\end{figure}

As shown in Figure~\ref{fig:ablations_reconstruction}, iterative subtraction strictly underperforms token masking in reconstructing the original weight vector. Subtraction captures significantly less of the cumulative explained norm (top row) and achieves lower overall cosine similarity with the original weight (bottom row) across iterations for both $W_{\text{gate}}$ and $W_{\text{out}}$. This suggests that geometrically projecting out the channel permanently degrades the weight vector's remaining latent structure, making subsequent feature extraction less effective. Token masking, by contrast, preserves the original geometry of $\mathbf{w}$ while successfully steering the kurtosis objective toward novel semantic directions.

\paragraph{Using more than 1 Householder matrix}
\label{appx:housholder}
A single Householder matrix ($k=1$) is technically a reflection rather than a proper rotation. Composing two Householder matrices ($k=2$) yields a true rotation. In practice, however, we find that a single reflection is entirely sufficient. As illustrated in Figure~\ref{fig:ablations_reconstruction}, the $k=2$ configuration performs virtually identically to the $k=1$ baseline across all metrics and weight types, with their curves overlapping almost perfectly. This confirms that a single reflection provides the necessary degrees of freedom to align the basis with high-kurtosis directions, rendering the added complexity and parameterization of multiple Householder matrices unnecessary.

\subsection{Hyperparameters selection}\label{appx:hyperparams}

Table~\ref{tab:hyperparams} summarizes the grid search results for our hyperparameter configurations. Hyperparameters were evaluated on a held-out set of 100 neurons per model/layer combination (disjoint from the experimental evaluation set) via grid search over the Cartesian product of: learning rate $\eta \in \{8\times10^{-4}, 2\times10^{-3}\}$, regularization coefficient $\lambda \in \{0.1, 0.3, 0.5\}$, and standard deviation threshold $\sigma \in \{4.0, 6.0, 8.0\}$. 

Because the metrics clustered heavily by the regularization penalty, we report the highest-performing configuration for each $\lambda$ value. Configurations were ranked by maximizing the harmonic mean of two metrics: 

First, \emph{orthogonality score} measures how mathematically distinct the discovered channel directions are from one another. It is defined as $1$ minus the mean absolute pairwise cosine similarity between all pairs of distinct extracted direction vectors $\mathbf{d}_i$ and $\mathbf{d}_j$:
\begin{equation}
    \text{Orthogonality Score} = 1 - \frac{1}{N(N-1)} \sum_{i \neq j} \frac{|\mathbf{d}_i \cdot \mathbf{d}_j|}{\lVert \mathbf{d}_i \rVert \lVert \mathbf{d}_j \rVert}
\end{equation}
where $N$ is the total number of channels. Taking the absolute value ensures that both highly correlated and highly anti-correlated directions are penalized. 

Second, \emph{explained norm} measures the proportion of the neuron's original magnitude that is captured by the learned channels. It is calculated as $1$ minus the relative reconstruction error:
\begin{equation}
    \text{Explained Norm} = 1 - \frac{\lVert \mathbf{w} - \hat{\mathbf{w}} \rVert}{\lVert \mathbf{w} \rVert}
\end{equation}
where $\mathbf{w}$ is the original neuron weight vector, $\hat{\mathbf{w}}$ is the reconstructed neuron vector, and $\lVert \mathbf{w} - \hat{\mathbf{w}} \rVert$ represents the $L_2$ norm of the reconstruction error (the residual). 

The number of optimization steps per channel was fixed at $n_\text{step} = 3000$.

\begin{table}[ht]
\centering
\small
\begin{tabular}{cccccc}
\toprule
\textbf{$\lambda$} & \textbf{Best $\eta$} & \textbf{Best $\sigma$} & \textbf{Explained Norm} & \textbf{Final Orthogonality} & \textbf{Harmonic Mean} \\
\midrule
\textbf{0.3} & $2\times10^{-3}$ & 4.0 & 0.72 & 0.78 & \textbf{0.749} \\
\textbf{0.5} & $8\times10^{-4}$ & 6.0 & 0.63 & 0.87 & 0.731 \\
\textbf{0.1} & $8\times10^{-4}$ & 4.0 & 0.86 & 0.54 & 0.663 \\
\bottomrule
\end{tabular}
\caption{Summary of hyperparameter grid search, reporting the best performing configuration (by harmonic mean) for each kurtosis regularization coefficient ($\lambda$). $\eta$: learning rate, $\sigma$: standard deviation threshold for token masking.}
\label{tab:hyperparams}
\end{table}

\subsection{Computational budget}\label{appx:compute}

\paragraph{Method efficiency}
\methodname{} operates entirely on model weights and requires no activation data, making its compute cost independent of dataset size. This contrasts sharply with activations-based baseline, which require collecting and processing millions of activation vectors before training can begin.

\paragraph{Parallelism and independence}
Each neuron's optimization is fully independent: the loss and gradient for a neuron depends only on its own rotation matrix and weight vector, with no coupling to other neurons. We exploit this structure by stacking all neurons in a chunk into a single batched tensor of shape $[\text{chunk size}, k, d_{\text{model}}]$ and running gradient descent on all of them in one forward--backward pass, with no interference between neurons. We use chunks of 5{,}000 neurons. One iteration (extracting one channel per neuron) takes approximately 11 minutes for a chunk of 5{,}000 neurons on a single H100 GPU.

\paragraph{Hardware and timing}
All experiments were run on a single NVIDIA H100 GPU. Applying \methodname{} to all neurons in one layer (extracting 50 channels per weight vector) takes approximately 3.8 GPU-hours for Gemma-2-2B-it (9,216 neurons per layer) and approximately 6.7 GPU-hours for Llama-3.1-8B-Instruct (14,336 neurons per layer). The 100-neuron experimental sample used for evaluation completes in under 30 minutes per layer.

\subsection{Vocabulary channels token-rank analysis}\label{appx:input_side}

The regularization term in Eq.~\ref{eq:loss} anchors each channel $\mathbf{v}_i$ to the original weight vector $\mathbf{w}$, achieving high reconstruction similarity (\S\ref{appx:reconstruction}). This raises a natural concern; the top tokens of a channel projection $\mathbf{v}_i\mathbf{U}$ are merely a re-ordering of the top tokens of the raw neuron projection $\mathbf{w}\mathbf{U}$.

To address this, we sample 20 gate neurons from Gemma-2-2B-it (10 each from layers 4 and 18) and, for every channel discovered by \methodname{}, compute the rank of each of the channel's top 50 tokens in the raw neuron projection $\mathbf{w}\mathbf{U}$. If channels simply re-ranked the neuron's top tokens, these ranks would concentrate at the top of the raw logit list.

We observe that across channels, $97.1\%$ of channel top tokens fall outside the top-200 tokens of the raw projection (rank $\geq 200$), with a median rank of $122{,}302$ out of $V = 256{,}000$. Vocabulary channels thus surface tokens that direct vocabulary projection of $\mathbf{w}$ never reaches, rather than re-ordering the tokens it already exposes.

\subsection{Limitations}\label{appx:limitations}

\methodname{} operates under a deliberate inductive bias: it searches for features that are aligned with the model's vocabulary. 
A significant body of work has identified functional components that operate in latent subspaces orthogonal to the vocabulary, such as confidence regulation mechanisms \citep{entropy-neuron} or positional processing features \citep{positional-neurons}. Such components fall outside the scope of our decomposition. Nevertheless, our completeness results (\S\ref{subsec:completeness}) demonstrate that vocabulary-aligned channels account for a substantial portion of neuron behavior, suggesting that this signal, while not exhaustive, still captures an accessible and significant layer of MLP computation.

In addition, we evaluate two layers per model across two architectures, selected based
on alignment to the vocabulary basis. Extending to additional scales, and architectures is a valuable next step.

\section{Qualitative examples}\label{appx:qualex}
In this section, we provide example channels obtained by \methodname{} (see Table~\ref{tab:channel-examples}) and analyze the interplay between $\mathbf{w}_{\text{gate}}$, $\mathbf{w}_{\text{in}}$, $\mathbf{w}_{\text{out}}$ channels within the gated MLP,
illustrating how vocabulary channels getting us closer to better understanding of the mechanisms behind neuron behavior.
We examine Neuron~9005 in Layer~18 of Gemma-2-2B-it
(Figure~\ref{fig:qualitative-neuron-9005}). This neuron activates positively on technical text involving negation and polarity concepts (e.g., comparison operators in C code, formal identities discussing \texttt{+} and \texttt{-}) and negatively on temporal deferral constructions (e.g., 
\emph{``it wasn't until 1817''}, 
\emph{``for many years''}).

\paragraph{Input side: when and why.}
\methodname{} explains this dual behavior through the interaction of gate and
value ($\mathbf{w}_{\text{in}}$) channels. On the positive side, $\mathbf{w}_{\text{gate}}$ channel~2
(\emph{``negative, Negative''}) detects contexts where negation or polarity is discussed, while $\mathbf{w}_{\text{in}}$ channel~1 (\emph{``negative, positive''}), a polarity concept signal aligns positively with the input ($\mathbf{w}_\text{in}{\cdot}\mathbf{x} = +1.76$). The product
$\sigma(\mathbf{w}_\text{gate}{\cdot}\mathbf{x}) \cdot
(\mathbf{w}_\text{in}{\cdot}\mathbf{x})$ is positive, yielding activation
$+2.76$. 
On the negative side, $\mathbf{w}_{\text{gate}}$ channel~0 \emph{``until, Until''}, detects temporal markers, while $\mathbf{w}_{\text{in}}$ channel~6 strongly anti-aligns with these inputs ($\mathbf{w}_\text{in}{\cdot}\mathbf{x} = -2.25$), producing activation $-4.53$.

\paragraph{Output side: what is promoted.}
The output-side channels complete the picture by revealing what the neuron writes
to the residual stream for each activation sign. Output channels discovered by
\methodname{} carry both kurtosis (sparsity) and skewness (directionality):
positive-skew channels have their semantically meaningful tokens on the positive
(promoted) side, while negative-skew channels have them on the negative
(suppressed) side. Since a negative neuron activation flips the sign of the
output contribution, negative-skew channels effectively have their bottom tokens
promoted when the neuron fires negatively.

Concretely, when the neuron fires positively, it promotes polarity vocabulary
through output channel~4 (\emph{``negative, positive''}, a polarity concept signal, aligns positively with skewness $= +4.6$),
along with code-closing syntax (ch\,1, skew $= +8.7$) and dashes (ch\,2, skew $= +6.3$). 
When the neuron fires negatively, the sign flip promotes the bottom
tokens of negative-skew channels: negation contractions \emph{``wasn't,
didn't, weren't''} (ch\,0, skew $= -4.1$), multilingual temporal markers
\emph{``until, Till, hasta, jusqu''} (ch\,3, skew $= -4.2$), and temporal delay vocabulary \emph{``wait, waiting''} (ch\,5, skew $= -4.2$).

This example demonstrates how vocabulary channels provide a much more nuanced and more mechanistic account: the input-side $\mathbf{w}_{\text{gate}}$ $\times$ $\mathbf{w}_{\text{in}}$ decomposition explains \emph{when} and \emph{why} the neuron activates with a particular sign, while the output-side channels, organized by skewness, explain \emph{what} the neuron promotes for each sign. 
Notably, the output channels reveal that this single neuron implements two coherent but distinct functions depending on activation polarity.
All channel are discovered entirely from weights, without any activation data.

\begin{figure*}[t]
\centering
\small
\renewcommand{\arraystretch}{1.15}

\begin{minipage}[t]{0.48\textwidth}
\centering
\textbf{\textcolor{posgreen}{Fires Positively (top examples)}}\\[3pt]
\begin{tabular}{@{}p{0.93\linewidth}@{}}
\toprule
\texttt{"2)) < (w2)) \&\& (((x1) - (x2)) \textbf{>} -(w1))"} \\
{\footnotesize Code with comparison/negation operators \hfill act.\ $= +2.76$} \\[2pt]
\texttt{"Operator x - y produces the same result as x \textbf{+} (-y)"} \\
{\footnotesize Formal text on positive/negative polarity \hfill act.\ $= +2.74$} \\
\bottomrule
\end{tabular}
\end{minipage}%
\hfill
\begin{minipage}[t]{0.48\textwidth}
\centering
\textbf{\textcolor{negred}{Fires Negatively (bottom examples)}}\\[3pt]
\begin{tabular}{@{}p{0.93\linewidth}@{}}
\toprule
\texttt{"Still, it wasn't \textbf{until} 1817 that the city..."} \\
{\footnotesize Temporal deferral construction \hfill act.\ $= -4.53$} \\[2pt]
\texttt{"...the utility and effectiveness \textbf{for} many years."} \\
{\footnotesize Temporal duration \hfill act.\ $= -3.49$} \\
\bottomrule
\end{tabular}
\end{minipage}


\noindent\textbf{Input Side: $\mathbf{w}_{\text{gate}}$ $\times$ $\mathbf{w}_{\text{in}}$ channel decomposition} \hfill \emph{(explains \textbf{when} and \textbf{why} the neuron fires $+/-$)}

\vspace{4pt}

\begin{minipage}[t]{0.48\textwidth}
\centering
\begin{tabular}{@{}p{0.93\linewidth}@{}}
\textbf{$\mathbf{w}_{\text{gate}}$} ch\,2: \emph{``negative, Negative''} ($\sigma = 2.16$) \\
{\footnotesize Detects contexts involving negation/polarity.} \\[4pt]
\textbf{$\mathbf{w}_{\text{in}}$} ch\,1: \emph{``negative, positive''} ($\mathbf{w}_\text{in}{\cdot}\mathbf{x} = {+}1.76$) \\
{\footnotesize Polarity concept signal (93\% of top examples).} \\
{\footnotesize \textcolor{posgreen}{Aligns} with input $\Rightarrow$ $\sigma(\cdot) \times (+) > 0$} \\[3pt]
\hfill \textbf{Predicted: $> 0$ \quad True: $+2.76$}
\end{tabular}
\end{minipage}%
\hfill
\begin{minipage}[t]{0.48\textwidth}
\centering
\begin{tabular}{@{}p{0.93\linewidth}@{}}
\textbf{$\mathbf{w}_{\text{gate}}$} ch\,0: \emph{``until, Until''} ($\sigma = 4.41$) \\
{\footnotesize Fires on temporal markers (100\% of bottom ex.).} \\[4pt]
\textbf{$\mathbf{w}_{\text{in}}$} ch\,6: \emph{``until, Until''} ($\mathbf{w}_\text{in}{\cdot}\mathbf{x} = {-}2.25$) \\
{\footnotesize Strongly \textcolor{negred}{anti-aligns} with temporal contexts.} \\
{\footnotesize $\sigma(\cdot) \times (-) < 0$} \\[3pt]
\hfill \textbf{Predicted: $< 0$ \quad True: $-4.53$}
\end{tabular}
\end{minipage}


\noindent\textbf{Output Side: Vocabulary channels with signed skewness} \hfill \emph{(explains \textbf{what} the neuron promotes)}

\vspace{4pt}

\begin{minipage}[t]{0.48\textwidth}
\centering
\textcolor{posgreen}{\textbf{Positive activation promotes} (positive-skew channels):}\\[3pt]
\begin{tabular}{@{}p{0.93\linewidth}@{}}
ch\,4 (skew $= +4.6$): \emph{``negative, positive, Negative''} \\
{\footnotesize Polarity vocabulary---the predicted concept.} \\[3pt]
ch\,1 (skew $= +8.7$): \texttt{']); "]); "));} \\
{\footnotesize Code closing syntax.} \\[3pt]
ch\,2 (skew $= +6.3$): \emph{``-{}-'', ``---'', ``—''} \\
{\footnotesize Minus sign, dashes and separators.} \\
\end{tabular}
\end{minipage}%
\hfill
\begin{minipage}[t]{0.48\textwidth}
\centering
\textcolor{negred}{\textbf{Negative activation promotes} (negative-skew, sign-flipped):}\\[3pt]
\begin{tabular}{@{}p{0.93\linewidth}@{}}
ch\,0 (skew $= -4.1$): \emph{``wasn't, weren't, didn't''} \\
{\footnotesize Negation contractions.} \\[3pt]
ch\,3 (skew $= -4.2$): \emph{``until, Till, hasta, jusqu''} \\
{\footnotesize Temporal markers (multilingual).} \\[3pt]
ch\,5 (skew $= -4.2$): \emph{``wait, waiting, waited''} \\
{\footnotesize Temporal waiting/delay.} \\
\end{tabular}
\end{minipage}

\vspace{8pt}

\caption{%
\textbf{Complete mechanistic decomposition of Neuron~9005 (Layer~18, Gemma-2-2B-it) via vocabulary channels.}
\textbf{Top:} The neuron activates positively on technical text with negation/polarity concepts and negatively on temporal deferral.
\textbf{Middle:} \methodname{}'s input-side $\mathbf{w}_{\text{gate}}$ and $\mathbf{w}_{\text{in}}$ channels explain the sign of the activation, the $\mathbf{w}_{\text{gate}}$ detects relevant context, while the $\mathbf{w}_{\text{in}}$ channel's alignment or anti-alignment with the input determines the sign.
\textbf{Bottom:} Output-side channels, organized by skewness sign, reveal what the neuron writes to the residual stream.
Positive activation promotes polarity vocabulary (\emph{``negative''}, \emph{``positive''});
negative activation promotes temporal negation tokens (\emph{``wasn't''}, \emph{``until''}, \emph{``wait''}).
All channels are discovered from weights alone.
}
\label{fig:qualitative-neuron-9005}
\end{figure*}

\begin{table}[t]
\centering
\small
\setlength{\tabcolsep}{2pt}
\begin{tabular}{@{}ll l c p{5.2cm} p{3.2cm}@{}}
\toprule
\textbf{Model} & \textbf{Neuron} & \textbf{MLP type} & \textbf{Ch} &
  \textbf{Top tokens} & \textbf{Description} \\
\midrule
\multirow{9}{*}{\shortstack[l]{Gemma-2\\-2b-it}}
  & \multirow{9}{*}{(18,\,6528)}
  & \multirow{3}{*}{$W_{\mathrm{gate}}$}
  & 0  & \textit{ride, Ride, riding, rides, ridden}           & Direct riding vocab. \\
  & & & 47 & \textit{platform, Platform, platforms}                    & Platform \\
  & & & {38} & \textit{school, School}                      & Dampens school ctx. \\
\cmidrule{3-6}
  & & \multirow{3}{*}{$W_{\mathrm{in}}$}
  & {0}  & \textit{ride, riding, rides, bike, horseback}   & Riding / locomotion \\
  & & & {16} & \textit{donkey, donkeys, horse, horses, mule} & Animals / mounts \\
  & & & 22         & \textit{gl, Gl, GL}                           & \texttt{gl-} subtoken \\
\cmidrule{3-6}
  & & \multirow{3}{*}{$W_{\mathrm{out}}$}
  & {0} & \textit{ride, riding, Ride, bike, motorcycle}    & Suppresses riding \\
  & & & 1          & \textit{mother, Mother, mom, father, parent}  & Promotes parenting \\
  & & & {9}  & \textit{mechanical, Mechanical, mechanism}    & Suppresses mechanics \\
\midrule
\multirow{9}{*}{\shortstack[l]{Llama-3.1\\-8B-Instruct}}
  & \multirow{9}{*}{(18,\,496)}
  & \multirow{3}{*}{$W_{\mathrm{gate}}$}
  & 0          & \textit{instruction, instructions, directions}  & Instructions \\
  & & & 2          & \textit{accept, Accept, acceptance}           & Acceptance \\
  & & & {7}  & \textit{charge, Charge, charges, fee}         & Dampens charges/fees \\
\cmidrule{3-6}
  & & \multirow{3}{*}{$W_{\mathrm{in}}$}
  & {0}  & \textit{instructions, directions}               & Instructions \\
  & & & {3}  & \textit{loyalty, loyal, faithful, allegiance} & Loyalty \\
  & & & {4}  & \textit{control, Control}                     & Control \\
\cmidrule{3-6}
  & & \multirow{3}{*}{$W_{\mathrm{out}}$}
  & 0          & \textit{follow, Follow}                         & Following \\
  & & & 6          & \textit{order, orders}                        & Orders \\
  & & & 7          & \textit{submission, submit, obedience}         & Submission \\
\bottomrule
\end{tabular}
\caption{%
  Selected vocabulary channels for two example neurons, across
  $W_{\mathrm{gate}}$, $W_{\mathrm{in}}$, and $W_{\mathrm{out}}$ weight matrices.
  Top tokens (up to 5) shown per channel.
}
\label{tab:channel-examples}
\end{table}

\section{Additional experimental details}
\label{appx:sec_details}

\subsection{Baselines}\label{appx:extra_baselines}

\paragraph{SAE-based baselines} Following \citet{gur-arieh-etal-2025-precise}, we disentangle MLP gate neurons using sparse autoencoders (SAEs) as a baseline for comparison with \methodname{}. We employ the Gemma Scope and Llama Scope SAEs \citep{gemmascope, llamascope}, which are trained on the residual stream at each neuron's respective layer. For each neuron, we take the top $k=15$ vectors from the SAE's out projection matrix with the highest dot product with said neuron, treating these vectors as the SAE-based counterpart to \methodname{}'s channels.

\paragraph{TopK-Iter}
We implement an iterative logit-lens decomposition: project $\mathbf{w}$ through $\mathbf{U}$, take the top $k=50$ tokens as a ``channel'', mask them, and repeat.
Channel descriptions are produced with the same pipeline as \methodname{} channels (\S\ref{subsec:exp_setup}) and evaluated with the identical faithfulness and completeness protocols. We evaluate on 20 neurons (10 per layer) from layers 4 and 18 of Gemma-2-2B-it. TopK-Iter achieves $0.10$ faithfulness (vs.\ \methodname{}'s $0.46$--$0.47$ on these layers) and $4\%$ completeness, well below the $20\%$ chance level (vs.\ $55$--$60\%$).

\paragraph{FastICA}
As a data-driven counterpart sharing \methodname{}'s kurtosis-deflation structure, we run FastICA \citep{fastica} (kurtosis contrast, orthogonal deflation) on the top and bottom token embeddings most similar to $\mathbf{w}$, treating the recovered components as channels under the same description and evaluation pipeline (20 neurons sampled from each of layers 4 and 18 of Gemma-2-2B-it).
FastICA performs at or below chance, faithfulness $0.25$, completeness $0.18$, with components converging to a few overlapping directions with \methodname{} but missing most of the channels carrying faithfulness and completeness.

Both weight-space alternatives fall far below \methodname{}: TopK-Iter reaches a faithfulness of only $0.10$ and a completeness of $4\%$, and FastICA performs at or below chance (faithfulness $0.25$, completeness $0.18$), despite sharing \methodname{}'s kurtosis-deflation structure (evaluated on Gemma-2 layers 4 and 18). Neither raw logit-lens inspection nor data-driven kurtosis deflation recovers the structure that \methodname{}'s weight-anchored rotations do.

\subsection{Input-side results}\label{appx:input_side}

\begin{figure}[ht]
    \centering
    \includegraphics[width=0.95\linewidth]{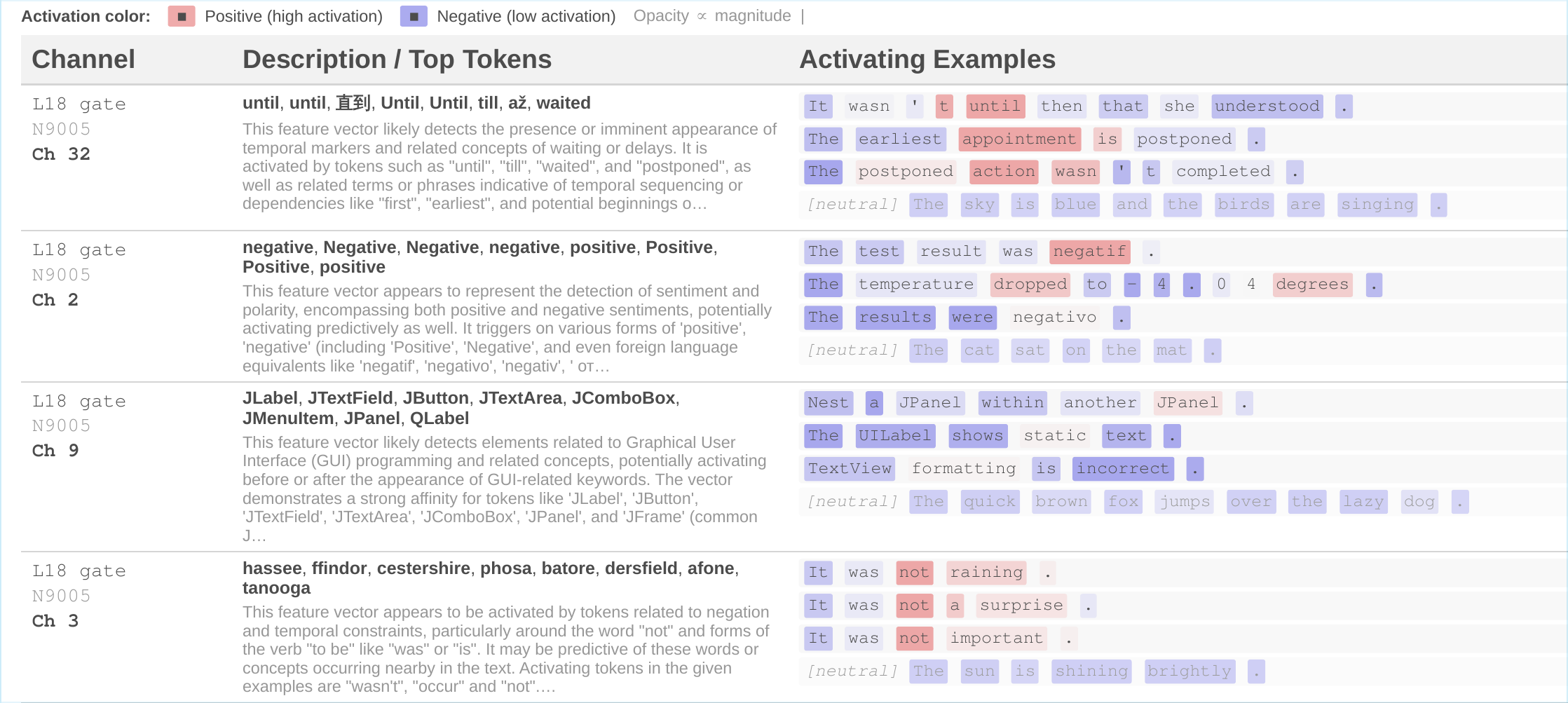}
    \caption{Visualization of four gate channels of Neuron~9005 (Layer~18, Gemma-2-2B-it).
    Each row shows a channel's top vocabulary tokens, its natural-language description, and
    three activating examples alongside one neutral example. Token color indicates activation
    polarity (red: positive, blue: negative) and opacity scales with magnitude. The channels
    capture distinct concepts: temporal markers, polarity/negation, GUI programming tokens,
    illustrating the fine-grained, interpretable
    structure recovered by \methodname{} from a single neuron's weight vector.}
    \label{fig:channel-viz}
\end{figure}

Figure~\ref{fig:channel-viz} illustrates four representative gate channels of Neuron~9005,
showing the top tokens, description, and activating examples for each.

\begin{figure}[ht]
    \centering
    \includegraphics[width=1.0\linewidth]{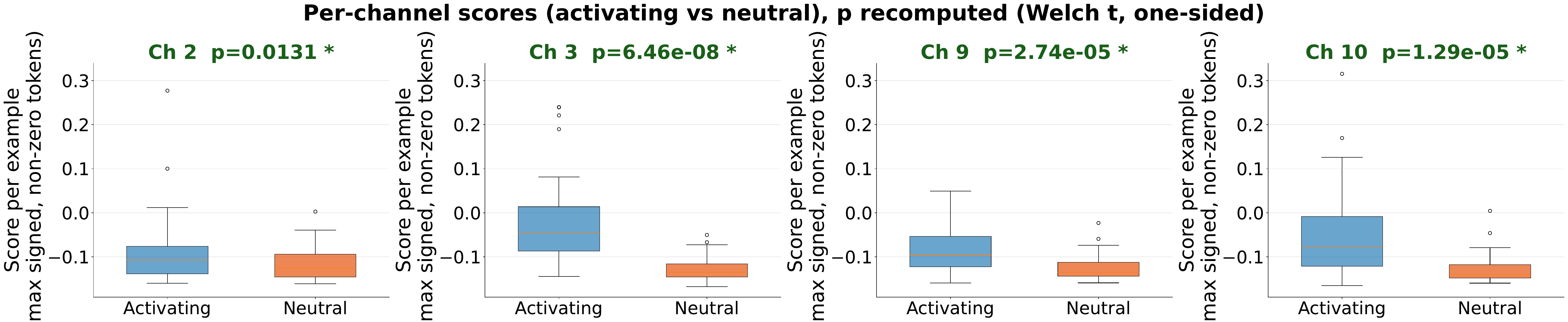}
    \caption{Per-channel faithfulness scores for representative gate channels of Neuron~9005
    (Layer~18, Gemma-2-2B-it). Each panel shows the distribution of peak neuron activations
    on activating (blue) vs.\ neutral (orange) sentences generated from the channel description.
    Channels shown all pass the one-sided t-test at $p < 0.05$ (indicated in the title
    of each panel), confirming that their descriptions reliably distinguish activating from
    non-activating inputs.}
    \label{fig:faithfulness-strip}
\end{figure}

Figure~\ref{fig:faithfulness-strip} shows the per-channel faithfulness results for the 4 gate
channels of Neuron~9005 (Layer~18, Gemma-2-2B-it).
For each channel, Gemini-2.0-Flash
generates 40 activating and 40 neutral sentences from the channel description; we compare peak neuron activations via a one-sided Welch t-test at $p < 0.05$. 
The four panels in Figure~\ref{fig:faithfulness-strip}
show representative passing channels, where activating sentences consistently elicit higher peak activations than neutral ones.

\paragraph{Activating / Neutral Example Generation Prompt}
Given a channel description, we prompt an LLM to generate synthetic sentences expected to activate the neuron (positive) and sentences that should not (negative), following the protocol described in \S\ref{subsec:input-side}. The full prompt is shown in Figure~\ref{fig:prompt_faithfulness}.

\subsection{Completeness setup}\label{appx:completeness}
For each gate weight vector we retrieve a random subset of 100 out of its top-1000 activating examples from
$\mathcal{D}$ and identify, for each example $\mathbf{x}$, the top channel
$\mathbf{v}^* = \arg\max_{\mathbf{v} \in \mathcal{C}}(\mathbf{x} \cdot \mathbf{v})$.
We then present an LLM judge (Gemini-3.1-Flash-Lite) with:
\begin{enumerate}[nosep,leftmargin=1.5em]
  \item The activating token context, with the highest-activating token marked
        \texttt{**like this**}.
  \item Five candidate descriptions: the description of $\mathbf{v}^*$ (correct)
        and four distractors drawn uniformly at random from channels of
        \emph{other} neurons in the same model and layer set.
\end{enumerate}
The judge selects the description it believes best explains why the neuron fired;
we record a hit when it selects the correct description.

\paragraph{Example.}
Below is a sample query for Neuron~9005 (Layer~18, Gemma-2-2B-it),
where the neuron fired on the token \texttt{**wasn't**}.

\begin{tcolorbox}[title=Sample Completeness Query --- Neuron 9005]
\small
\textbf{Sentence:}
``It   \texttt{**wasn't**}   what I expected at all.''

\textbf{Candidate descriptions:}
\begin{enumerate}[nosep,leftmargin=1.5em]
  \item Riding and locomotion contexts (horses, bikes, vehicles).
  \item Polarity/negation constructions: contractions like \textit wasn't, didn't, can't.
        \hfill\textbf{[correct]}
  \item Instruction-following and obedience vocabulary.
  \item Technical programming and software development tokens.
  \item Temporal markers indicating future scheduling.
\end{enumerate}
\textbf{Judge response:} ``The sentence contains `wasn't', a negation contraction.
Description~2 best matches.''
\end{tcolorbox}

\noindent
The four distractor descriptions are sampled from random neurons in Gemma Layer~18.
In this example the judge selects Description~2, the correct vocabulary channel.

\subsection{Patchscopes setup}\label{appx:patchscopes}

We use the Patchscopes framework \citep{patchscopes} to decode semantic content
encoded in a neuron's output weight vector $\mathbf{w}_{\text{out}}$.
We construct the few-shot prompt
\[
\underbrace{%
  \texttt{cat}\to\texttt{cat};\;\texttt{135}\to\texttt{135};\;
  \texttt{hello}\to\texttt{hello};}_{\text{few-shot context}}
\quad\texttt{?}
\]
where the \texttt{?} probe token's residual-stream representation (at the input to
block~0) is overwritten with the scaled weight vector $\alpha\,\mathbf{w}_{\text{out}}$
before the forward pass continues.
The few-shot context biases the model to ``read'' the semantic content of the
injected vector rather than predicting from syntactic context alone.

\paragraph{Why scaling by $\alpha$ is necessary.}
Token embeddings in Gemma-2-2B-it have $\ell_2$ norm on the order of
$\|\mathbf{e}_t\| \approx 100$--$150$,
whereas MLP output weight vectors have norm $\|\mathbf{w}_{\text{out}}\| \approx 0.5$--$2$.
Injecting the raw weight vector ($\alpha=1$) therefore places the probe far outside
the distribution of token embeddings, yielding near-degenerate generations.
Multiplying by $\alpha$ rescales the probe into the normal embedding range:
\[
  \mathbf{p}_\alpha = \alpha\,\mathbf{w}_{\text{out}}.
\]
We sweep $\alpha \in \{-400,-350,\ldots,350\}$ (step 50).
Setting $\alpha > 0$ amplifies the semantic content of $\mathbf{w}_{\text{out}}$;
setting $\alpha < 0$ probes its \emph{semantic opposite} by flipping the injected
direction, which for a dual-polarity neuron surfaces the other polarity cluster.

\paragraph{Channel ablation.}
To test the causal role of a specific channel $\mathbf{v}$, we ablate it from
$\mathbf{w}_{\text{out}}$ before injecting:
\[
  \mathbf{w}_{\text{ablated}}
    = \mathbf{w}_{\text{out}}
      - \frac{\mathbf{w}_{\text{out}} \cdot \mathbf{v}}{\|\mathbf{w}_{\text{out}}\|^2}\,\mathbf{v},
\]
where $\mathbf{v}$ is the channel vector (not unit-normalised).
The weight $(\mathbf{w}_{\text{out}}\cdot\mathbf{v})/\|\mathbf{w}_{\text{out}}\|^2$
measures how much of $\mathbf{w}_{\text{out}}$'s length is contributed by $\mathbf{v}$.
We then inject $\alpha\,\mathbf{w}_{\text{ablated}}$ and compare the decoded output
to the baseline injection $\alpha\,\mathbf{w}_{\text{out}}$ at $\alpha=400$.

\paragraph{Decoding parameters.}
We run 20 independent sampling passes for each alpha value of the baseline and 10 for each alpha value ablated variant (temperature~$=0.9$, up to 8 new tokens per pass).
All generated tokens are pooled into a single multi-set per condition.

\paragraph{Metric.}
Let $T_\mathbf{v}$ be the top-50 vocabulary-projection tokens of channel $\mathbf{v}$.
Define the concept-token fraction for a weight vector $\mathbf{w}$ as
\[
f(\mathbf{w})
  = \frac{\bigl|\{t \in \operatorname{pool}(\mathbf{w}) : t \in T_\mathbf{v}\}\bigr|}
         {|\operatorname{pool}(\mathbf{w})|}.
\]
The relative change when channel $\mathbf{v}$ is ablated is
\[
\Delta
  = \frac{f(\mathbf{w}_{\text{ablated}}) - f(\mathbf{w}_{\text{out}})}
         {f(\mathbf{w}_{\text{out}})}
  \in (-1,\,1),
\qquad
\mathbf{w}_{\text{ablated}}
  = \mathbf{w}_{\text{out}} - (\mathbf{w}_{\text{out}} \cdot \mathbf{v})\,\mathbf{v}.
\]
\emph{Self-channel ablation} monitors the fraction of $T_\mathbf{v}$ tokens when
$\mathbf{v}$ itself is ablated;
\emph{cross-channel ablation} monitors the same fraction when a different channel
$\mathbf{v}' \neq \mathbf{v}$ is ablated instead.
A faithful, non-redundant channel should produce $\Delta_{\text{self}} \approx -1$
and $\Delta_{\text{cross}} \approx 0$.

\paragraph{Example.}
For out-channel~0 of Neuron~9005
(top tokens: \textit{wasn't, weren't, didn't, can't, isn't}),
self-ablation reduces the fraction of polarity tokens from
${\approx}18\%$ to ${\approx}2\%$ ($\Delta \approx -89\%$),
while cross-ablation of an unrelated channel leaves it near $18\%$
($\Delta \approx +15\%$).

\subsection{LLM judge validation}\label{appx:judge_validation}
Two evaluation tasks in this paper rely on LLM judges: completeness (\S\ref{subsec:completeness}), judged by Gemini-3.1-Flash-Lite, and head-to-head description comparison (\S\ref{sec:enhancing}), judged by Gemini-3-Flash.
We use different judges as the completeness task is simpler and requires substantially more LLM calls, making a lightweight model preferable.
To assess whether these LLM judges are reliable substitutes for human annotators (NLP graduate students), we apply the Alternative Annotator Test~\citep{llm-judge}, which tests whether an LLM can statistically replace a human annotator within an annotation group.
For each task, three annotators independently annotated 50 instances following the same protocols as the LLM judge. For the head-to-head task, description order was randomized and annotators were blind to method identity. We set $\varepsilon = 0.15$, which is suited for skilled annotators, and a $p{-}value = 0.05$.

On the completeness task , Gemini-3.1-Flash-Lite achieves $\bar{\rho}_f = 0.89$ (vs.\ $\bar{\rho}_h = 0.81$ for humans), with $\omega = 2/3$.
On the head-to-head task , Gemini-3-Flash achieves $\bar{\rho}_f = 0.897$ vs.\ $\bar{\rho}_h = 0.885$, with $\omega = 3/3$.
Both tasks pass the $\omega \geq 0.5$ threshold, confirming that the LLM judges can reliably substitute for human annotation in these comparative evaluation settings.

\section{Additional Details on Neuron Description Generation}\label{appx:description_generation}

\subsection{Variant Selection via Pairwise Evaluation}\label{appx:variant-selection}

\paragraph{Vocab-channel aggregation strategies}
We experimented with four strategies for aggregating the 25 gate and 25 $\mathbf{w}_{\text{in}}$ channel descriptions into a single per-polarity neuron description.
The variants differ in (a) which gate channels are included and (b) how $\mathbf{w}_{\text{in}}$ channels are filtered by skewness polarity.
Table~\ref{tab:vc_variants} summarizes the four strategies.

\begin{table}[h]
\centering
\small
\begin{tabular}{lll}
\toprule
\textbf{Variant} & \textbf{$\mathbf{w}_{\text{gate}}$ channels} & \textbf{$\mathbf{w}_{\text{in}}$ channels} \\
\midrule
All gate, all in                & all                  & all \\
Positive-skew gate, all in      & positive-skew only   & all \\
All gate, positive-skew in      & all                  & positive-skew only \\
All gate, negative-skew in      & all                  & negative-skew only \\
\bottomrule
\end{tabular}
\caption{Four aggregation strategies for \methodname{} neuron descriptions. The last two variants separate positive and negative activation regimes by filtering $\mathbf{w}_{\text{in}}$ channels according to the sign of their vocabulary-projection skewness, while retaining all $\mathbf{w}_{\text{gate}}$ channels.}
\label{tab:vc_variants}
\end{table}

\paragraph{MaxAct baseline variants}
We evaluated three versions of the MaxAct+VocabProj baseline, differing in what information is provided to the LLM:
\textbf{v1}: top-20 activating examples only (one combined description);
\textbf{v2} (selected): top-20 examples concatenated with the top-50 vocabulary tokens from the $\mathbf{w}_{\text{in}}$ and $\mathbf{w}_{\text{gate}}$ vector projections, producing polarity-split descriptions;
\textbf{v3}: same as v2 but with $\mathbf{w}_{\text{in}}$ and $\mathbf{w}_{\text{gate}}$ vocabulary projections described separately before synthesis.

\paragraph{Stage 1 evaluation}
To select the best variant within each method, we ran pairwise LLM-judged comparisons (Gemini-2.0-Flash) across all variants, separately for positive- and negative-polarity activation contexts.
We used 20 randomly sampled neurons from Llama-3.1-8B-Instruct, with 50 examples per neuron sampled from the top-1000 Pile activations.
Position bias was controlled by running each comparison twice with swapped description order and declaring a winner only when both orderings agree.
Table~\ref{tab:stage1_variants} reports the win rates.

\begin{table}[h]
\centering
\small
\begin{tabular}{llll}
\toprule
\textbf{Method} & \textbf{Polarity} & \textbf{Variant} & \textbf{Win rate} \\
\midrule
\multirow{4}{*}{\methodname{}} & \multirow{4}{*}{positive}
  & \texttt{all\_gate\_split\_positive} & \textbf{78.3\%} \\
& & \texttt{all\_gate\_all\_in}         & 35.0\% \\
& & \texttt{positive\_gate\_all\_in}    & 31.7\% \\
\midrule
\multirow{3}{*}{\methodname{}} & \multirow{3}{*}{negative}
  & \texttt{all\_gate\_split\_negative} & \textbf{57.5\%} \\
& & \texttt{all\_gate\_all\_in}         & 37.5\% \\
\midrule
\multirow{3}{*}{MaxAct+VocabProj} & \multirow{3}{*}{positive}
  & v2                                  & \textbf{67.5\%} \\
& & v1                                  & 57.5\% \\
& & v3                                  & 25.0\% \\
\midrule
\multirow{3}{*}{MaxAct+VocabProj} & \multirow{3}{*}{negative}
  & v2                                  & \textbf{47.1\%} \\
& & v1                                  & 61.8\% \\
& & v3                                  & 40.6\% \\
\bottomrule
\end{tabular}
\caption{Stage 1 within-method variant win rates on 20 neurons from Llama-3.1-8B-Instruct.
Bold denotes the selected variant for each method and polarity.
For \methodname{}, we select \texttt{all\_gate\_split\_positive} (positive) and \texttt{all\_gate\_split\_negative} (negative).
For MaxAct+VocabProj, we select v2 for  as it enriches the activation-based evidence with vocabulary-projection tokens from both  $\mathbf{w}_{\text{gate}}$ and $\mathbf{w}_{\text{in}}$, providing the baseline with the strongest available signal and ensuring the most competitive comparison against \methodname{}.}
\label{tab:stage1_variants}
\end{table}

This section details the full prompting pipeline used in \S\ref{sec:enhancing}.
 
\subsection{Channel-level description}\label{appx:multi-channel-desc}
Each of the 25 $\mathbf{w}_{\text{gate}}$ and 25 $\mathbf{w}_{\text{in}}$ channels is independently described  by prompting an LLM with the channel's top-50 vocabulary tokens and up to 5 top-activating examples. The full prompt is shown in Figure~\ref{fig:prompt_channel_desc}.
 
\subsection{Neuron-level synthesis (polarity-split)}\label{appx:channel-synth}
The individual channel descriptions are then synthesized into a single neuron description, separately for positive and negative activations. $\mathbf{w}_{\text{gate}}$ and $\mathbf{w}_{\text{in}}$ channel descriptions are provided together, organized by role. The full prompt is shown in Figure~\ref{fig:prompt_synthesis}.
 
\paragraph{Baseline: MaxAct+VocabProj description}
For the MaxAct+VocabProj baseline, we prompt the LLM with 20 top-activating examples and the top/bottom-50 vocabulary tokens from the $\mathbf{w}_{\text{gate}}$ and $\mathbf{w}_{\text{in}}$ weight vector projections. The full prompt is shown in Figure~\ref{fig:prompt_maxact}.
 
\paragraph{Head-to-head pairwise evaluation}
For the LLM-judged pairwise comparison described in \S\ref{sec:enhancing}, each comparison is run twice with swapped description order; a winner is declared only when both orderings agree. The full judge prompt is shown in Figure~\ref{fig:prompt_h2h}.

\subsection{Head-to-head examples }\label{appx:wins-losses}
Table~\ref{tab:win} presents selected head-to-head comparisons between \methodname{}'s unified neuron descriptions and those produced by the MaxAct++ and MaxAct+VocabProj baselines. For each neuron, we show the descriptions generated by all three methods alongside a representative activating example from the Pile positive split. The final column indicates whether the LLM judge preferred the \methodname{} description for that example. These cases illustrate how \methodname{}'s vocabulary-grounded decomposition often yields more specific and faithful descriptions, particularly for neurons encoding structured or syntactic patterns that activation-based methods tend to summarize in overly generic terms.

\begin{table}[h]
\centering
\resizebox{\textwidth}{!}{%
\begin{tabular}{l p{4.0cm} p{7.5cm} p{5.0cm} c}
\toprule
\textbf{Layer, Neuron} & \textbf{Activating example}& \textbf{ROTATE description} & \textbf{Baseline description} &\textbf{Win/Loss} \\
\midrule
L22, N6946 & 
\texttt{/// Get the host name associated with the entry. template <class Allocator> std\textbf{*::*}basic\_string, std::char\_traits<char>, Allocator> host\_name( const Allocator\& alloc}   

\textbf{Pile top [100-500]} & 
This neuron activates on contexts related to sleep, rest, and altered states of consciousness (dreaming, falling asleep), alongside concepts of returning or restarting, often involving function words (\emph{to, of, you}) and morphological elements. Additionally, it responds to notions of bursting/failure, central locations/functions, and suspension/hanging, and \textbf{code snippets related to filtering operations on arrays}. & 
This neuron activates on words related to sleep, sleeping, snoring, and waking up, as well as general personal pronouns and common function words like ``to'', ``or'', and ``of'', possibly reflecting awareness of narrative context involving sleep. 

[\textbf{MaxAct+VocabProj}]&
Win \\


\midrule
L22, N1939 &
\texttt{"thumbnail",
                "file",
                "fanart",
                "streamdetails"
            ],
            "\textbf{*player*}id": 1
        ],
        "id": "VideoGetItem"
Check this out\}
} 

\textbf{Pile top [0-100]}
&
This neuron activates in contexts blending organizational systems, financial elements, and technical details, particularly those involving data processing and structured information. This includes: pipelines and routing of data, archives and architecture, financial assets and payments, macro/micro scale comparisons, lists/catalogs, letters/alphabets, notes/records, and measurements of volume. It is also sensitive to names and identifiers, particularly those containing the letter sequence 'ee' &
This neuron activates on code snippets, particularly related to the VLC media player library (libVLC) or JSON-\textbf{RPC calls for media players} (like XBMC), often involving player control methods. It also activates on articles, 'the' and 'to'
[\textbf{MaxAct+VocabProj}]&
Loss\\
\midrule
L12, N496 &
We just cruised on her to the Panama Canal last week! The Maitre'De in\textbf{* the*} Posh Dining Room Goran Gorigjewski is awesome!!

\textbf{Pile top [0-100]}
&
This neuron activates positively in contexts involving the definite \textbf{article 'the' alongside varied semantic themes} including: workplace interactions; self-reference; code overrides; strength/resilience; sending/transmission; philosophical concepts/proper nouns; authentication ('login'); geographical locations/cardinal directions; physical actions; and potentially female names. This suggests an emphasis on contextually defined entities within narrative or technical contexts
&
proper nouns; context indicating inquiry or explanation

\textbf{[MaxAct++]} &
Win\\
\midrule

L18, N2241 &
The English prose \textbf{*poem*} is a verse form that is usually unrhymed and written in the...

\textbf{FineWeb top [0-100]}

 &
This neuron strongly activates on code snippets, configurations, and technical documentation, often featuring specific numerical identifiers, compound words, and elements related to authorship or provenance. It also demonstrates sensitivity to partial words and specific syllables ('an', 'on', 'ol', 'ug', 'ac') and common suffixes. Addition-related terms, Slavic language fragments, and spoiler/coupon contexts can also trigger activation. &
references to \textbf{ poetic forms}, styles, or innovation 

\textbf{[MaxAct++]} &
Loss\\

\bottomrule
\end{tabular}%
}
\caption{Example wins and losses of \methodname{} in head-to-head comparisons against MaxAct++ and MaxAct+VocabProj.}
\label{tab:win}
\end{table}
 
\begin{figure*}[t]
\begin{tcolorbox}[title=Polarity-Split Neuron Description Synthesis Prompt]
\small
You are analyzing a neuron in a language model. Below are descriptions of individual channels that correspond to \texttt{\{polarity\}} activations of the neuron. Try to make this compressed as you can, but still touch on the diverse features of the neuron. A description should be no more than 50 words.
 
Layer: \texttt{\{layer\_idx\}}\quad Neuron: \texttt{\{neuron\_idx\}}\quad Polarity: \texttt{\{polarity\}}
 
\textbf{$\mathbf{w}_{\text{gate}}$ Channel Descriptions} (control what activates the neuron):\\
\texttt{\{gate\_descriptions\}}
 
\textbf{$\mathbf{w}_{\text{in}}$ Channel Descriptions} (determine the neuron's activation --- \texttt{\{polarity\}} activation):\\
\texttt{\{in\_descriptions\}}
 
Your task is to synthesize these channel descriptions into a single, coherent description of what causes \texttt{\{polarity\}} activations of this neuron.
 
Create a unified description that:
\begin{enumerate}[nosep,leftmargin=1.5em]
\item Identifies the common semantic or syntactic themes across channels
\item Explains what inputs activate this neuron
\item Notes any patterns in token appearance vs prediction
\item Is specific enough to be useful but general enough to capture the neuron's overall function
\end{enumerate}
 
\textbf{Avoid vague descriptions:} Do NOT use generic, uninformative descriptions like ``diverse set of linguistic and semantic features'' or ``various textual patterns''. Be SPECIFIC about what causes \texttt{\{polarity\}} activations. If the channels are truly diverse, list the 2--3 most prominent specific patterns rather than using vague umbrella terms.
 
Please return your answer in JSON format.
\end{tcolorbox}
\caption{Polarity-split neuron description synthesis prompt (\S\ref{sec:enhancing}). $\mathbf{w}_{\text{gate}}$ and $\mathbf{w}_{\text{in}}$ channel descriptions are provided separately; the LLM produces a unified description of at most 50 words. Used with Gemini-2.0-Flash.}
\label{fig:prompt_synthesis}
\end{figure*}
 
\begin{figure*}[t]
\begin{tcolorbox}[title=MaxAct+VocabProj Baseline Description Prompt]
\small
You are analyzing a neuron in a language model.
 
Layer: \texttt{\{layer\_idx\}}\quad Neuron: \texttt{\{neuron\_idx\}}\quad Analysis Type: \texttt{\{polarity\_upper\}} ACTIVATIONS
 
Below are the \texttt{\{polarity\_description\}} activating examples for this neuron (gate $\times$ in activation). The activating token in each example is marked with \texttt{**asterisks**}. The number in brackets \texttt{[X.XX]} is the activation value.
 
\textbf{\texttt{\{polarity\_upper\}} Activating Examples:}
\texttt{\{examples\}}
 
Additionally, here is the LogitLens analysis showing which tokens the neuron's gate and in vectors project to in vocabulary space:
 
\textbf{$\mathbf{w}_{\text{gate}}$ Vector Projection:}\\
--- Top tokens (most similar in vocabulary space): \texttt{\{gate\_top\_tokens\}}\\
--- Bottom tokens (least similar): \texttt{\{gate\_bottom\_tokens\}}
 
\textbf{$\mathbf{w}_{\text{in}}$ Vector Projection:}\\
--- Top tokens: \texttt{\{in\_top\_tokens\}}\\
--- Bottom tokens: \texttt{\{in\_bottom\_tokens\}}
 
Your task is to analyze these examples and LogitLens projections to describe what causes \texttt{\{polarity\}} activations. The description should be no more than 50 words.
 
Consider:
\begin{enumerate}[nosep,leftmargin=1.5em]
\item What semantic or syntactic patterns appear in these \texttt{\{polarity\}}-activation examples?
\item How do the LogitLens tokens relate to \texttt{\{polarity\}} activation patterns?
\item Is there a coherent theme?
\end{enumerate}
 
\textbf{Avoid vague descriptions:} Do NOT use generic, uninformative descriptions like ``diverse set of linguistic and semantic features'' or ``various textual patterns''. Be SPECIFIC. If the examples are diverse, list the 2--3 most prominent specific patterns rather than using vague umbrella terms.
 
Please return your answer in JSON format.

\end{tcolorbox}
\caption{MaxAct+VocabProj baseline description prompt (\S\ref{sec:enhancing}). Combines 20 top-activating examples with LogitLens vocabulary projections of the $\mathbf{w}_{\text{gate}}$ and $\mathbf{w}_{\text{in}}$ vectors. Used with Gemini-2.0-Flash.}
\label{fig:prompt_maxact}
\end{figure*}
 
\begin{figure*}[t]
\begin{tcolorbox}[title=Head-to-head pairwise evaluation]
\small
You are comparing two neuron descriptions to determine which more accurately explains an activating pattern.
 
Below is a sequence of tokens that highly activates a specific neuron. Tokens with high activation values are marked with their activation in brackets.
 
\textbf{Activating Sequence:}
\texttt{\{formatted\_tokens\}}
 
\textbf{Description A:}
\texttt{\{description\_a\}}
 
\textbf{Description B:}
\texttt{\{description\_b\}}
 
\textbf{Task:}
Determine which description more accurately identifies the specific pattern that causes this neuron to activate on the highlighted tokens.
 
Important guidelines:
\begin{itemize}[nosep,leftmargin=1.5em]
\item Focus on ACCURACY, not on level of detail or length. A short, precise description can be better than a long, vague one.
\item Do NOT prefer a description just because it is longer or more detailed.
\item A description may cover multiple themes. It should win if at least one of its themes correctly explains the highlighted example. Do not penalize a description for covering themes beyond what appears in this example.
\item Choose ``TIE'' if both descriptions capture the activation pattern equally well, even if one is more detailed.
\end{itemize}
 
Choices: ``A'' if Description A more accurately identifies the activation pattern; ``B'' if Description B does; ``TIE'' if both are equally accurate (or equally inaccurate).
 
\textbf{Response Format (JSON only):}\\
\texttt{\{\{"winner": "A" or "B" or "TIE",}\\
\texttt{~~"reasoning": "<one sentence explanation>"\}\}}
\end{tcolorbox}
\caption{Head-to-head pairwise evaluation prompt (\S\ref{sec:enhancing}). Each comparison is run twice with swapped order; a winner is declared only when both orderings agree. Used with Gemini-3-Flash.}
\label{fig:prompt_h2h}
\end{figure*}

\section{Prompts used in experiments}\label{appx:prompts}
\paragraph{Channel description}\label{appx:channel-desc}

Each channel is described by prompting an LLM with the channel's top-50 vocabulary tokens and up to 5 top-activating examples. The full prompt is shown in Figure~\ref{fig:prompt_channel_desc}.

\begin{figure*}[t]
\begin{tcolorbox}[title=Channel Description Prompt]
\small
I am analyzing a channel (component) of a neuron in a language model.
 
Associated tokens (sorted by logit value):
\texttt{\{tokens\_str\}} 

Additionally, here are real text examples where this channel activates strongly:
\texttt{\{examples\_section\}}
 
Task:
\begin{enumerate}[nosep,leftmargin=1.5em]
\item Identify the common semantic or syntactic theme among these tokens and examples.
\item Provide a short description of what this channel likely represents or detects.
\item The description should be specific but capture the general concept.
\end{enumerate}
 
Please return your answer in JSON format.\\
\end{tcolorbox}
\caption{Prompt used to describe a single vocabulary channel. Each channel is described independently before synthesis into a neuron-level description. Used with Gemini-2.0-Flash.}
\label{fig:prompt_channel_desc}
\end{figure*}

\paragraph{Activating / neutral example generation prompt}
Given a channel description, we prompt an LLM to generate synthetic sentences expected to activate the neuron (positive) and sentences that should not (negative), following the protocol described in \S\ref{subsec:input-side}. The full prompt is shown in Figure~\ref{fig:prompt_faithfulness}.

\begin{figure*}[t]
\begin{tcolorbox}[title=Activating / neutral example generation prompt]
\small
I'm going to give you explanations and interpretations of features from LLMs. You must take in each explanation, and generate \texttt{\{num\_positive\}} sentences for which you think the feature will have a high activation, and \texttt{\{num\_negative\}} for which they'll have a low activation.
 
\textbf{For the high activation examples:}
\begin{itemize}[nosep,leftmargin=1.5em]
\item Make sure to choose ones that will cause a high activation with high confidence
\item You don't have to include all groups, just make examples that you're confident will have high activation
\item Make the sentences both include words from the explanation AND represent the concept
\item Try to use specific examples and make them literal interpretations of the explanation, without trying to generalize
\end{itemize}
 
\textbf{For the low activation examples:}
\begin{itemize}[nosep,leftmargin=1.5em]
\item These should have nothing to do with the interpretation
\item They should be orthogonal and completely unrelated to the feature
\end{itemize}
 
\textbf{Output Format:}
You must output strictly valid JSON with the following structure:\\
\texttt{\{\{"max\_activation": ["example 1", "example 2", ...],}\\
\texttt{~~"min\_activation": ["example 1", "example 2", ...]\}\}}
 
Explanation: \texttt{\{explanation\}}
\end{tcolorbox}
\caption{Prompt used to generate activating and neutral examples for the input-side faithfulness evaluation (\S\ref{subsec:input-side}). Default: 40 positive + 40 negative examples. Used with Gemini-2.0-Flash.}
\label{fig:prompt_faithfulness}
\end{figure*}

\paragraph{Completeness LLM judge prompt}
The 5-way channel matching prompt used for the completeness evaluation is shown in Figure~\ref{fig:prompt_completeness}.
 
\begin{figure*}[h]
\begin{tcolorbox}[title=5-Way Channel Matching Prompt (Completeness)]
\small
You are going to be given a sentence, and five descriptions of different components in a language model. One of these descriptions describes a component that was highly activated when processing the sentence, while the other four descriptions are unrelated. The most highly activated token in the sentence is marked with double asterisks (\texttt{**like this**}). Your task is to identify this description. Please respond with a short line describing your reasoning, and then the second line should contain only the number (1--5) corresponding to the correct description. Please respond exactly in this format. Even if you are unsure, make your best guess.
 
\textbf{Sentence:}
``\texttt{\{activating\_string\}}''
 
\textbf{Descriptions:}
\texttt{\{chan\_descs\}}
\end{tcolorbox}
\caption{5-way channel matching prompt used for the completeness evaluation (\S\ref{subsec:completeness}). The LLM judge (Gemini-3.1-Flash-Lite) selects which of five candidate descriptions best matches the activating input.}
\label{fig:prompt_completeness}
\end{figure*}


\end{document}